\documentclass[format=acmsmall, review=false]{acmart}

\pdfoutput=1 

\renewcommand\footnotetextcopyrightpermission[1]{} 

\setcopyright{none}

\usepackage{booktabs} 
\usepackage{dsfont}  
\usepackage[ruled]{algorithm}  
\usepackage{makecell} 
\usepackage{xifthen} 
\usepackage{tikz}
 \usepackage{csquotes} 
\usepackage{aliascnt} 
\usepackage[capitalise,noabbrev]{cleveref}
\usepackage{algpseudocode} 
\algnewcommand\algorithmicparameters{\textbf{Parameters:}}
\algnewcommand\Parameters{\item[\algorithmicparameters]}

\theoremstyle{acmplain}

\newtheorem{hyptest}{Hypothesis Test}

\newcommand{\R}{{\rm I\!R}}
\newcommand{\dataset}{\textit{Data}}
\newcommand{\ndim}{d}
\newcommand{\ngenu}{n}
\newcommand{\nart}{n_\textit{art}}
\newcommand{\datasetext}{\dataset_\textit{Ext}}
\newcommand{\inst}[1][]{%
	\ifthenelse{\isempty{#1}}{\textit{inst}}{\textit{inst}_{#1}}%
}
\newcommand{\instval}[1][]{%
	\ifthenelse{\isempty{#1}}{\textit{inst}}{\textit{inst}^{(#1)}}%
}
\newcommand{\neigh}[1][]{%
	\ifthenelse{\isempty{#1}}{\textit{neigh}}{\textit{neigh}_{#1}}%
}
\newcommand{\neighset}{\textit{Neighs}}

\newcommand{\artinst}[1][]{%
	\ifthenelse{\isempty{#1}}{\textit{art}}{\textit{art}_{#1}}%
}
\newcommand{\artinstset}{\textit{Arts}}
\newcommand{\artout}[1][]{%
	\ifthenelse{\isempty{#1}}{\textit{out}}{\textit{out}_{#1}}%
}
\newcommand{\artoutset}{\textit{ArtOuts}}
\newcommand{\genuoutset}{\textit{Outs}}
\newcommand{\rand}[1][]{%
	\ifthenelse{\isempty{#1}}{\textit{rand}}{\textit{rand}_{#1}}%
}
\newcommand{\randval}[1][]{%
	\ifthenelse{\isempty{#1}}{\textit{rand}}{\textit{rand}^{(#1)}}%
}

\newcommand{\outdist}[1][]{%
	\ifthenelse{\isempty{#1}}
	{
	\textit{Out}\!\left(\cdot\right)
	}{
	\textit{Out}\!\left(#1\right)
	}%
}
\newcommand{\outdistgen}[2][]{%
	\ifthenelse{\isempty{#1}}
	{
		\textit{Out}_{#2}\!\left(\cdot\right)
	}{
		\textit{Out}_{#2}\!\left(#1\right)
	}%
}
\newcommand{\outdistgenu}[1]{\outdistgen[#1]{\textit{Genu}}}
\newcommand{\normdist}[1][]{%
	\ifthenelse{\isempty{#1}}
	{
		\textit{Norm}\!\left(\cdot\right)
	}{
		\textit{Norm}\!\left(#1\right)
	}%
}
\newcommand{\normdistgenu}[1][]{%
	\ifthenelse{\isempty{#1}}
	{
		\textit{Norm}_{\textit{Genu}}\!\left(\cdot\right)
	}{
		\textit{Norm}_{\textit{Genu}}\!\left(#1\right)
	}%
}
\newcommand{\normdistonenoarg}[1]{\textit{Norm}_{\textit{Genu}}^{(#1)}\!\left( \cdot \right)}
\newcommand{\normdistone}[1]{\textit{Norm}_{\textit{Genu}}^{(#1)}\!\left( \inst^{(#1)} \right)}

\newcommand{\approachFont}[1]{\textsc{#1}}
\newcommand{\unifBox}{\approachFont{unifBox}}
\newcommand{\lhs}{\approachFont{lhs}}
\newcommand{\unifSphere}{\approachFont{unifSphere}}
\newcommand{\maniSamp}{\approachFont{maniSamp}}
\newcommand{\marginSample}{\approachFont{marginSample}}
\newcommand{\distBased}{\approachFont{distBased}}
\newcommand{\boundVal}{\approachFont{boundVal}}
\newcommand{\invHist}{\approachFont{invHist}}
\newcommand{\infeasExam}{\approachFont{infeasExam}}
\newcommand{\densAprox}{\approachFont{densAprox}}
\newcommand{\surReg}{\approachFont{surReg}}
\newcommand{\skewBased}{\approachFont{skewBased}}
\newcommand{\boundPlace}{\approachFont{boundPlace}}
\newcommand{\negSelect}{\approachFont{negSelect}}
\newcommand{\negShift}{\approachFont{negShift}}
\newcommand{\gaussTail}{\approachFont{gaussTail}}
\newcommand{\ganGen}{\approachFont{ganGen}}

\newcommand{\classifier}{\mathcal{C}}
\newcommand{\trainGen}{\textit{trainGen}}
\newcommand{\testGen}{\textit{testOuts}}
\newcommand{\trueOuts}{\textit{trueOuts}}
\newcommand{\artOuts}{\textit{artOuts}}

\newcommand{\classifierset}{\textit{Classifiers}}
\newcommand{\design}{\textit{Design}}
\newcommand{\inrset}{\textit{Intr}\artoutset}
\newcommand{\Ex}[2]{\mathbb{E}_{#1}\!\left[ #2 \right]}
\newcommand{\indicator}[1]{\mathds{1}_{\left( #1 \right)}}

\begin{document}
	\usetikzlibrary{arrows,shapes,positioning,shadows,trees}

	\fancyfoot{}

	\title[Generating Artificial Outliers in the Absence of Genuine Ones --- a Survey]{Generating Artificial Outliers \\in the Absence of Genuine Ones --- a Survey}

	\author{Georg Steinbuss}
	\email{georg.steinbuss@kit.edu}
	\orcid{0000-0002-2051-3394}
	\affiliation{%
		\institution{Karlsruhe Institute of Technology (KIT)}
		\city{Karlsruhe}
		\country{Germany}
	}
	\author{Klemens B\"{o}hm}
	\email{klemens.boehm@kit.edu}
	\affiliation{%
		\institution{Karlsruhe Institute of Technology (KIT)}
		\city{Karlsruhe}
		\country{Germany}
	}
	
	\begin{abstract}
		By definition, outliers are rarely observed in reality, making them difficult to detect or analyse. 
		Artificial outliers approximate such genuine outliers and can, for instance, help with the detection of genuine outliers or with benchmarking outlier-detection algorithms. 
		The literature features different approaches to generate artificial outliers. 
		However, systematic comparison of these approaches remains absent. 
		This surveys and compares these approaches.
		We start by clarifying the terminology in the field, which varies from publication to publication, and we propose a general problem formulation. 
		Our description of the connection of generating outliers to other research fields like experimental design or generative models frames the field of artificial outliers. 
		Along with offering a concise description, we group the approaches by their general concepts and how they make use of genuine instances. 
		An extensive experimental study reveals the differences between the generation approaches when ultimately being used for outlier detection.
		This survey shows that the existing approaches already cover a wide range of concepts underlying the generation, but also that the field still has potential for further development. 
		Our experimental study does confirm the expectation that the quality of the generation approaches varies widely, for example, in terms of the data set they are used on. 
		Ultimately, to guide the choice of the generation approach in a specific context, we propose an appropriate general-decision process.
		In summary, this survey comprises, describes, and connects all relevant work regarding the generation of artificial outliers and may serve as a basis to guide further research in the
		field.
	\end{abstract}

	
	
	\maketitle
	\thispagestyle{empty}
	
\section{Introduction}

Outliers are data instances that deviate from normal ones~\cite{Hodge2004-zt,Chandola2009-fv,Theiler2003-dy,Fan2004-oq,Steinwart2005-vz}. 
Since they are abnormal, one is unlikely to observe them. 
In addition, given some notion of \enquote{normal}, outliers can deviate from that normal instances in infinite ways. Given their rarity and variety, developing methods to detect outliers is difficult. Nevertheless, numerous approaches to this identification exist~\cite{Hodge2004-zt,Chandola2009-fv}. 
If there are known outliers available for study, one can use classical supervised approaches for outlier detection (i.e., solve a \emph{very} imbalanced classification problem) \cite{Chandola2009-fv}. 
When such outliers are unavailable, many approaches have been developed to identify outliers in that circumstance as well. 
However, a learning task without a single instance from one of the classes of interest is difficult: because it is nearly impossible to evaluate the performance of different approaches for example. 
Thus, there exists a large body of literature on generating artificial outliers~\cite{Steinbuss2017-hv,Wang2018-vt,Curry2009-bu,Gonzalez2002-bf,Shi2006-jh,Steinwart2005-vz,Theiler2003-dy,Pham2014-lk,Wang2009-gg,Fan2004-oq,Abe2006-ca,Hempstalk2008-tu,Neugebauer2016-od,Desir2013-xp,Banhalmi2007-pa,Tax2001-na,Hastie2009-mu}. 
The idea is that one extends the given data set through accurate approximations of outliers and thus resolves the problem of an unknown class. 
Approaches to generate artificial outliers can rely on genuine outliers (i.e., outliers that have been observed and are not artificial). 
One famous approach of this kind is the Synthetic Minority Over-sampling Technique (SMOTE)~\cite{Chawla2002-ec}. 
However, the problem SMOTE tries to solve, differs from that of generating outliers without any genuine outliers available. 
In this study, we focus on approaches for outlier generation without genuine outliers.

\subsection{Purpose of This Survey}

The common ground for different uses of artificial outliers or approaches to generate such outliers remains unclear, mainly due to a limited general perspective. 
That is, what artificial outliers are used for in general and how existing approaches to generate them differ are currently not well formulated.
The absence of a sophisticated general perspective makes it also difficult to connect the generation of artificial outliers to other research fields, such as generative modelling or adversarial learning.
This integration, however, would be beneficial for both the generation of artificial outliers and for related fields.
One obstacle to such a general perspective, however, is that the terminology used in articles from different fields varies widely.

Possibly due to the missing general perspective, there is not much knowledge available on the performance of generating outliers or methods using them.
For example, we are aware of only one comparison of the two most common uses for artificial outliers: (1) casting an unsupervised learning task into a supervised one and (2) parameter tuning of one-class classifiers (see \cref{sec:use_of_arts} for details).
Both these uses for artificial outliers result in a method to detect real outliers.
In~\cite{Davenport2006-ck} the two uses are compared, but only for a few rather similar generation approaches.
Hence, we find it somewhat difficult to assess whether one of the two uses yields better outlier detection, irrespective of the generation approach used. 
A sizable study is also needed to investigate the hypothesis that a high-quality result using a specific generation approach is not general. 
In other words, other generation approaches might be better on, say, other data sets.

\subsection{Goals of This Survey}

With this survey we want to give the field of artificial outliers the missing general perspective.
This is, clarifying the differences of the many diverse approaches to generate artificial outliers that already exist but also formulating and discussing a more general problem formulation.

Having some general perspective we also aim at a sizeable study that features systematic comparisons in terms of uses and generation approaches for artificial outliers. 
In particular, we want to compare (1) the performance of the different generation approaches, (2) the difference in outlier-detection performance of the two common uses for artificial outliers, and (3) analyse the characteristics of the data (e.g., the number of attributes) that influence the performance of approaches for generating artificial outliers or using them. 
Another goal we have is to construct a concise set of advises to guide anyone in the application of artificial outliers.
These should simplify the usage of artificial outliers by much and thus might further increase their usage in the detection of real outliers.

\subsection{Methods}

We start this survey by establishing a unified terminology around artificial outliers.
We then describe the different usages of artificial outliers.
Following this, we highlight connections to other research fields and possible synergies. 
Given these connections, we produce a general problem formulation for the generation of artificial outliers and embed existing approaches into it. 
We describe each existing approach, using the unified terminology.
All this together results in the general perspective on the field of artificial outliers we aim at. 

We then perform extensive experiments, comparing the two most common uses for artificial outliers.
These also allow us to analyse the performance of the different generation approaches with many benchmark-data sets on outlier detection.
The effect of data characteristics like the number of attributes can be analysed as well.
Following the careful analysis of the results of our experiments we synthesize the findings obtained into a straightforward decision process that guides in the usage of artificial outliers. 

\subsection{Organization of This Survey}
    
The remainder of this article is structured as follows. 
We introduce a general terminology in \cref{sec:terminology}, and describe the usages of artificial outliers in \cref{sec:use_of_arts}.
We then establish connections between the topic of generating artificial outliers with other research fields in \cref{sec:con_to_other}.
In \cref{sec:problem}, we offer a general problem formulation. 
\cref{sec:place_approach} describes the different generation approaches that presently exist.
\cref{sec:filter} outlines methods to filter artificial outliers for ones that give better results than the set of unfiltered ones. \cref{sec:experiments} contains the results of our extensive experimental study, and \cref{sec:conclusion} presents our conclusions.

\section{Terminology and Notion} 
\label{sec:terminology}

In this section, we specify the terminology  used in this survey. We start by discussing terms that are ambiguous in the literature and proceed with further terminology and notions.

\subsection{Ambiguities in the Literature}

Certain issues arise in the process of describing a data set.  
\enquote{Instances} are also referred to as \enquote{examples}~\cite{Abe2006-ca,Banhalmi2007-pa,Curry2009-bu}, \enquote{objects}~\cite{Tax2001-na,Theiler2003-dy,Wang2009-gg}, \enquote{observations}~\cite{Steinwart2005-vz,Shi2006-jh}, \enquote{vectors}~\cite{Gonzalez2002-bf}, \enquote{input/sample}~\cite{lee2018-tr}, \enquote{data}~\cite{Wang2018-vt}, or \enquote{data points}~\cite{dai2017good}. 
Here we prefer the term \enquote{instances} throughout. 
Another issue is the naming of the different characteristics of instances. 
Common terms are \enquote{attributes}~\cite{Theiler2003-dy,Banhalmi2007-pa,Hempstalk2008-tu,Curry2009-bu,Wang2009-gg,Desir2013-xp,Steinbuss2017-hv}, \enquote{features}~\cite{Gonzalez2002-bf,Fan2004-oq,Steinwart2005-vz,Pham2014-lk,Neugebauer2016-od} or \enquote{dimensions}~\cite{Tax2001-na,Wang2018-vt}. 
We use \enquote{attribute}. 
Another ambiguity is the term for the set of all possible instances. 
For example, when the data set consists of $\ndim$ real valued attributes, the set of all possible instances is some subset of $\R^d$. 
Possible terms are \enquote{feature space}~\cite{Tax2001-na,Wang2009-gg,Neugebauer2016-od}, \enquote{space}~\cite{Gonzalez2002-bf}, \enquote{domain}~\cite{Fan2004-oq}, \enquote{input space}~\cite{dai2017good} or \enquote{region}~\cite{Steinbuss2017-hv}. 
We use \enquote{instance space}. 
	
There also are ambiguities in the general field of outlier detection. 
Most central is the notion of outliers itself. 
Aside from  \enquote{outlier}~\cite{Tax2001-na,Abe2006-ca,Hempstalk2008-tu,Curry2009-bu,Wang2009-gg,Desir2013-xp,Neugebauer2016-od,Steinbuss2017-hv,Wang2018-vt}, some authors use \enquote{anomaly}~\cite{Gonzalez2002-bf,Theiler2003-dy,Fan2004-oq,Steinwart2005-vz}, \enquote{out-of-distribution sample}~\cite{lee2018-tr}, \enquote{negative example}~\cite{Gonzalez2002-bf,Banhalmi2007-pa}, \enquote{counter example}~\cite{Banhalmi2007-pa}, \enquote{attack example}~\cite{Pham2014-lk} or \enquote{infeasible example}~\cite{Neugebauer2016-od}. 
We use \enquote{outlier}. 
The term for the counterpart of outliers is ambiguous as well. 
While they often are referred to as \enquote{normal} instances~\cite{Gonzalez2002-bf,Theiler2003-dy,Fan2004-oq,Steinwart2005-vz,Abe2006-ca,Hempstalk2008-tu,Pham2014-lk,Wang2018-vt}, other terms used include \enquote{inlier}~\cite{Steinbuss2017-hv}, \enquote{positive} instance~\cite{Banhalmi2007-pa} or \enquote{feasible} instance~\cite{Neugebauer2016-od}. 
We use \enquote{normal} instances. 
	
We see two notions with ambiguous terminology related to artificial outliers. 
These outliers often are referred to as \enquote{artificial} outliers~\cite{Tax2001-na,Theiler2003-dy,Fan2004-oq,Steinwart2005-vz,Abe2006-ca,Hempstalk2008-tu,Curry2009-bu,Wang2009-gg,Desir2013-xp,Pham2014-lk,Neugebauer2016-od}, and the procedure that creates them is referred to as \enquote{generation} \cite{Tax2001-na,Gonzalez2002-bf,Fan2004-oq,Steinwart2005-vz,Abe2006-ca,Shi2006-jh,Banhalmi2007-pa,Hempstalk2008-tu,Curry2009-bu,Wang2009-gg,Desir2013-xp,Pham2014-lk,Neugebauer2016-od,Wang2018-vt}. 
However, \citet{Wang2018-vt} use \enquote{pseudo}, and \citet{Shi2006-jh} use \enquote{synthetic} instead of \enquote{artificial}.
Instead of \enquote{generated}, \cite{Steinbuss2017-hv} uses \enquote{placed}, and \cite{lee2018-tr,dai2017good} use \enquote{sample}.
We will use \enquote{artificial} and \enquote{generated}.

\subsection{Further Terminology and Notation}
		
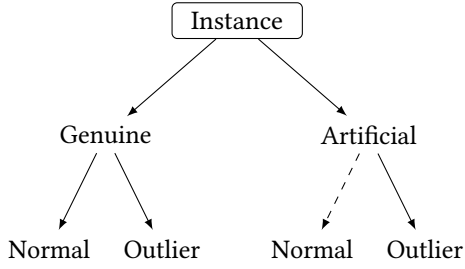
\begin{figure}[ht]
	\centering
	\begin{tikzpicture}[
	basic/.style  = {draw, text width=1.5cm, rectangle},
	root/.style   = {basic, rounded corners=2pt, thin, align=center},
	level 1/.style={sibling distance=35mm},
	level 2/.style={sibling distance=15mm},
	edge from parent/.style={->,draw},
	>=latex]
	
	\node[root] {Instance}
	child {node[level 1] (c1) {Genuine}
		child {node[level 2] (c11) {Normal}}
		child {node[level 2] (c12) {Outlier}}
	}
	child {node[level 1] (c2) {Artificial}
		child[dashed] {node[level 2] (c21) {Normal}}
		child {node[level 2] (c22) {Outlier}}
	};
	\end{tikzpicture}
	\caption{Terminology regarding instances}
	\Description{A tree that clarifies the four types of instances we have in the article. An instance can either be genuine or artificial and either normal or an outlier.}
	\label{fig:instance_names}
\end{figure}

In this survey we refer to four types of instances.
Instances are either genuine or artificial, each of which can be termed either normal or outlier (see \cref{fig:instance_names}).
In line with the majority of generating approaches, when we refer to genuine instances, we mean both normal and outlier instances that are not generated. 
In this survey, we focus on artificial outliers. 
Thus, artificial instances are outliers unless explicitly stated otherwise.

\cref{tab:notion_overview} summarizes our mathematical notation that we will detail in the following. 
Let the given data set $\dataset \in \R^{\ngenu \times \ndim}$ be a matrix in which each row represents a data instance and each column an attribute. 
Hence, we have $\ngenu$ instances with $\ndim$ attributes. 
Let $\inst$ denote an instance of any type from \cref{fig:instance_names} and $\artinst$, an artificial one. 
The value of the $i$th attribute of an instance $\inst$ is $\instval[i]$, where $i$ as a subscript refers to the ith value from a set. 
For example, $\inst[i]$ is the $i$th instance from $\dataset$. 
This notation generalizes to other objects (like the distributions from \cref{def:Inst_dist}) and also sets other than the data set.
Variable $l$ refers to the label of an instance (\emph{normal} or \emph{outlier}) in \cref{sec:gen_models}.
A set of artificial outliers is referred to as $\artoutset$, and one of genuine outliers, as $\genuoutset$. 
An \emph{interesting} set of artificial outliers (cf.\ \cref{def:interesting_outs}) is abbreviated with $\inrset$. 
A data set that is extended with $\nart$ artificial outliers is referred to as $\datasetext \in \R^{\ngenu+\nart \times \ndim}$, and $\classifier$ is the shorthand for a classifier, while $p(\cdot)$ denotes a probability density or mass function. 
It is a density function if the random variable to which it refers is continuous, and a mass function if this random variable is discrete.
The $k$-nearest neighbors of an instance are denoted as $\neighset$,
while a single nearest neighbor is denoted as $\neigh$.
Certain highly specific notions (e.g., the parameters of a generation approach) are not featured here. 
These are introduced where needed.
	
\begin{table}[ht]
	\centering
	\caption{Overview of our notions.} 
	\label{tab:notion_overview}
	\begin{tabular}{lr} 
		\toprule
		Notion & Meaning \\ 
		\midrule 
		$\dataset$ & Given data set \\
		$\datasetext$ & Data set with artificial outliers \\
		$\ndim$ & Number of data attributes \\
		$\ngenu$ & Number of genuine instances \\
		$\nart$ & Number of artificial outliers \\
		$\genuoutset$ & Set of genuine outliers \\ 
		$\artoutset$ & Set of artificial outliers \\
		$\inst$ & Any instance \\
		$l$ & Label of $\inst$ \\
		$\instval[i]$ & Value of $\inst$'s $i$th attribute \\
		$\inst[i]$ & $i$th instance from a set \\
		$\artinst$ & Artificial instance \\
		$\inrset$ & Set of interesting artificial outliers \\
		$\neighset$ & Set of $k$-nearest neighbors \\
		$\neigh$ &  Any neighbor\\
		$\classifierset$ & Set of classifers \\
		$\classifier$ & Any classifer\\
		$p(\cdot)$ & Probability density/mass function \\
		$\normdist$ & Distribution of normal instances \\
		$\outdist$ & Distribution of outliers \\
		\bottomrule
	\end{tabular}
\end{table}
    	
To conclude this section, \cref{def:Inst_dist} gives the notation for different distributions.
    
 \begin{definition}[$\outdist, \ \normdist$]
 	\label{def:Inst_dist}
 	$\outdist$ and $\normdist$ are the distributions of any instance $\inst$, $\outdist$ for outlier instances and $\normdist$ for normal ones. 
\end{definition}
A subscript indicates if the distribution is for a special type of instance. For example, $\normdistgenu{}$ denotes the distribution of normal genuine instances.

\section{Usage of Artificial Outliers}
\label{sec:use_of_arts}

Having the unified terminology from \cref{sec:terminology}, we now describe problems that one can solve with artificial outliers, subsequently referred to as \enquote{use cases}.
The joint description of these use case is the first building block for our general perspective on artificial outliers.

We are aware of three use cases from the literature: (1) casting an unsupervised learning task into a supervised one~\cite{Gonzalez2002-bf,Shi2006-jh,Steinwart2005-vz,Theiler2003-dy,Pham2014-lk,Fan2004-oq,Abe2006-ca,Hempstalk2008-tu,Neugebauer2016-od,Desir2013-xp,Banhalmi2007-pa,Hastie2009-mu,El-Yaniv2007-dv}, subsequently referred to as \enquote{casting task}; (2) parameter tuning of one-class classifiers~\cite{Wang2018-vt,Wang2009-gg,Tax2001-na,dai2017good}, referred to as \enquote{one-class tuning}; and (3) exploring properties of a specific type of outlier with artificial outliers~\cite{Steinbuss2017-hv}, referred to as \enquote{exploratory usage}.
We order the detailed description of each use case in the following by their relevance in the literature.

\subsection{Casting Task}

This use case is based on the observation that a data set extended with artificial outliers consists of two fully labelled classes: genuine and artificial instances. 
The genuine instances are those that have actually been observed, while artificial instances have been generated. 
Thus, one can apply any classifier to set the two apart. 
Genuine instances mostly are normal, and the artificial ones have been generated so that they are outliers. 
The classifier thus learns to distinguish between normal and outlying instances. 
Next, the number of artificial outliers is controllable. 
Thus, unlike \enquote{classical} supervised outlier detection, this classification does not even have to be unbalanced.

One reason this approach is common might be that it has a theoretical basis \cite{Hastie2009-mu,Hempstalk2008-tu,Abe2006-ca,Theiler2003-dy,Steinwart2005-vz,El-Yaniv2007-dv}. 
Given some data with unknown distribution, one can use a classifier that distinguishes genuine from artificial instances, to obtain a density estimation of the genuine instances. 
This in turn allows identifying instances that are unlikely. 
Section 14.2.4 (“Unsupervised as Supervised Learning”) in~\cite{Hastie2009-mu} and~\cite{Steinwart2005-vz} show this for different types of classifiers. 

\subsection{One-Class Tuning}
\label{sec:descp_one_class_optim}

Another use case is hyper-parameter tuning for one-class classifiers \cite{Wang2018-vt,Wang2009-gg,Tax2001-na}. 
The training of a one-class classifier uses only instances from one class to learn to separate new instances belonging to this class from those that do not~\cite{Hempstalk2008-tu}. 
Instances not belonging to the class are deemed \enquote{outliers}. 
A common one-class classifier belongs to the category of Support Vector Machines (SVMs): the Support Vector Data Description (SVDD) introduced in~\cite{Tax1999-mn}. 
It has hyperparameters $s$ and $\nu$~\cite{Tax2001-na} where $s$ is the kernel width, and $\nu$ is an upper bound on the fraction of genuine instances classified as outlying. 
To choose values for both parameters, one must optimize the error rate of the resulting one-class classifier~\cite{Tax2001-na}. 
However, since one-class classification is applied when there is either no outliers or not a sufficient number of outliers, estimating this error is difficult. 
Various approaches for the generation of artificial outliers have been developed to estimate the error~\cite{Wang2018-vt,Wang2009-gg,Tax2001-na}. 

While the two use cases described so far differ, their outcome is the same: a classifier for outlier detection. 
In both use cases, the artificial outliers help train the classifier. 
A good generation approach yields a high detection rate on outliers, be they genuine or artificial. 
To investigate the quality differences in terms of outlier detection between the two cases, we have performed experiments, see \cref{sec:compare_class}. 
We have found that there are some differences, but none of the two use cases is clearly preferable in terms of detection quality.

\subsection{Exploratory Usage} 
\label{sec:explore_task}

So-called \enquote{hidden outliers} are the object of study in \cite{Steinbuss2017-hv}. 
A hidden outlier is one that is detectable only in certain subsets of the attributes~\cite{Muller2012-cg,Steinbuss2017-hv}. 
Hidden outliers are not visible for detection schemes not explicitly looking at these subsets of attributes. 
Hence, they depict blind spots of the detection scheme that could be very dangerous for the system monitored. 
\cite{Steinbuss2017-hv} has derived properties of hidden outliers. 
An example is how dependencies among attributes or the number of genuine instances influence the occurrence of hidden outliers. 
\cite{Steinbuss2017-hv} uses artificial hidden outliers to study such characteristics. 
To this end, artificial outliers are generated and then filtered for hidden ones. 
One can then check how many artificial outliers are actually hidden. 
This methodology allows one to infer the characteristics of the data set and of the attribute subsets which influence the occurrence of hidden outliers. 
With these characteristics, one can develop methods to search for attribute subsets robust to hidden outliers. 
An attribute subset is robust to hidden outliers if generating or finding hidden outliers in it is difficult. 
Put generally, one can use artificial outliers to explore and analyse special kinds of outliers.

\section{Connection to Other Fields}
\label{sec:con_to_other}

The goals of the use cases for artificial outliers given in \cref{sec:use_of_arts} allow us now to connect the generation of artificial outliers to other research fields.
This is another building block for our general perspective on artificial outliers.

We see at least three broad research fields closely connected to artificial outliers: generative models, design of experiments and adversarial machine learning. 
The first two are fields from statistics, while the last one is a relatively new paradigm mostly from computer science.
In the following we will discuss the general ideas of each of these research fields but also their connection to the generation of artificial outliers.

\subsection{Generative Models}
\label{sec:gen_models}

The following discussion is mostly based on the work of Bernardo et al.~\cite{Bernardo2007-kh} on the connection between discriminative and generative models. 
In machine learning, one often tries to predict a label $l_i$ that belongs to an instance $\inst[i]$. In the remainder of this section, $l_i$ identifies $\inst[i]$ as \emph{normal} or \emph{outlier} (classification). 
The goal then is to determine the conditional probability $p(l \mid \inst)$ from a given data set $\dataset$ (i.e., the distribution of $l$ given an instance $\inst[i]$).
Two common approaches to do so are discriminative or generative, respectively. 
Discriminative models directly approximate $p(l \mid \inst)$, while generative ones first try to find the joint distribution $p(l, \ \inst)$. 
By sampling from this joint distribution, it is possible to generate instances. 
Hence, these models are called \enquote{generative}. 
Specifying the joint distribution $p(l, \ \inst)$ is usually done by defining a distribution for the classes $p(l)$ and a class-conditional distribution for the instances $p(\inst \mid l)$, along with finding the best fit to the instances in $\dataset$. 
This specification gives the joint distribution by
\begin{equation}
	p(l, \ \inst) = p(\inst \mid l) \cdot p(l).
\end{equation}
We have omitted the distribution parameters that are fitted using $\dataset$ for the sake of clarity. 

Since $l$ can only take two distinct values, the generative model is fully specified if $p(\inst \mid l = \textit{normal})$, $p(l=\textit{normal}) =: p_\textit{normal}$, $p(\inst \mid l = \textit{outlier})$ and $p(l=\textit{outlier}) =: p_\textit{outlier}$ are specified. 
Artificial outliers are essentially samples from $p(\inst \mid l = \textit{outlier})$ or at least approximations of these samples. 
To generate the artificial outliers, one explicitly or implicitly defines $p(\inst \mid l = \textit{outlier})$. 
With the number of samples generated,  $p_\textit{normal}$ and $p_\textit{outlier}$ are defined as well. 
Thus, when generating artificial outliers, most parts of the generative model are also defined. 
The only missing part is the distribution of normal instances $p(\inst \mid l = \textit{normal})$. 
Hence, if we explicitly define $p(\inst \mid l = \textit{outlier})$ and estimate $p(\inst \mid l = \textit{normal})$ from the data, we end up with a generative model for outlier detection. 
This, however, is not the only connection between artificial outliers and generative models. 
A generative model can also be used to classify instances as outlier or normal. 
This classification is also what artificial outliers facilitate in the use cases \emph{casting task} and \emph{one-class tuning}. Interestingly, outliers do not need to be generated for the generative model, since their distribution only needs to be defined. 
The issue with such an approach, however, is that estimating $p(\inst \mid l = \textit{normal})$ is not simple. 
The generation of outliers is often simpler. 
Thus, using some artificial outliers to train or tune a classifier is simpler or sometimes simply more effective than is specifying the generative model.
The connection of artificial outliers and generative models is strong.
If it is simple to, for instance, estimate $p(\inst \mid l = \textit{normal})$ in some setting, one might prefer the generative model over artificial outliers.

Another insight in this context comes from $p_\textit{outlier}$. 
We find it surprising that many inventors of generation approaches do not discuss its importance. 
Since $p_\textit{outlier}$ is part of the generative model, it clearly does affect the decision of whether an instance is an outlier or not.
Recall that in the case of artificial outliers, $p_\textit{outlier}$ is essentially given by $\nart$. 
Hence, $\nart$ also determines whether an instance is an outlier or not.
	
\subsection{Design of Experiments}
\label{sec:design_of__experiments}

The following description is based on \cite{Lovric2011-ql}. 

The \enquote{Design of Experiments} deals with modeling the dependence of a random variable $l$ on some deterministic factors $\instval[1], \dots , \, \instval[\ndim]$ (i.e., attribute values). 
A combination of the $\ndim$ deterministic factors yields an artificial instance $\inst$.
As in the previous section, $l$  identifies $\inst$ as \emph{normal} or \emph{outlier}.
The topic \enquote{Design of Experiments} aims to find a set of such factor combinations $\design = \{ \inst[1], \dots, \inst[\nart] \}$ that give \emph{optimal} results regarding $l$. 
To illustrate, \enquote{optimal} can mean that our classification with regard to $l$ yields a perfect accuracy. 
One does not need to estimate this classification from $\design$ alone. 
It is also reasonable  to consider that it is learned from $\datasetext = \design \ \cup \ \dataset$, like in the \emph{casting task} use case. 
Hence, the generation of artificial outliers can be seen as a subfield of the design of experiments. 
Although it is difficult to make the definition of \enquote{optimal} more concrete, we approach this in \cref{sec:problem}. 
The design of experiments encompasses extensive theoretical work. 
We believe that establishing a rigid connection of artificial outliers to this broad field may facilitate a rather formal derivation of relevant concepts and approaches.

To our knowledge, no previous work has been done regarding artificial outliers in the field of design of experiments. 
However, some rather general approaches to a good $\design$ seem to be applicable. 
One such approach is already common when generating artificial outliers ($\unifBox$, see \cref{sec:unifBox}) \cite{Lovric2011-ql}. 
It relies entirely on random sampling. 
This reliance makes it difficult to ensure that the whole instance space (e.g., $\R^{\ngenu \times \ndim}$) is evenly covered. However, such behavior often is a desirable property, since it is usually not known \emph{a priori} which regions of the instance space have to be covered. 
The Latin hypercube design ensures that the instances are evenly spread in the instance space \cite{Santner2013-at}. 
See \cref{def:lhs}. 

\begin{figure}[ht]
	\centering
	\includegraphics[width=0.35\linewidth]{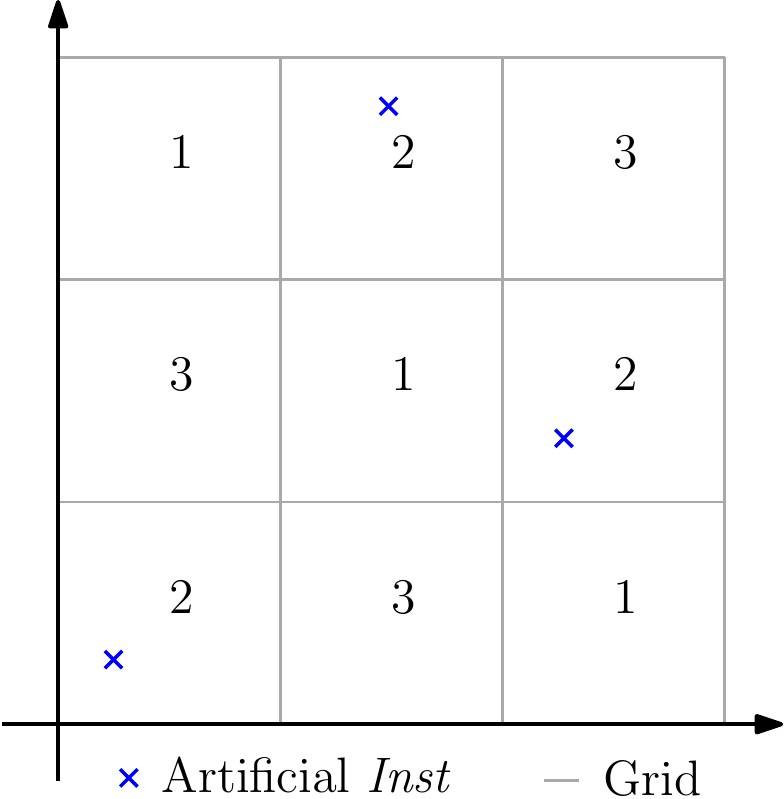}
	\caption{Illustration of the generation approach $\lhs$. $\nart = 3$ and $i= 2$.}
	\Description{An illustration of the Latin hypercube design. See Definition 3.1 for details.}
	\label{fig:examp_lhs}
\end{figure}

\begin{definition}[$\lhs$]
	\label{def:lhs}
	$\lhs$ is an approach to generate artificial instances using the so-called \emph{Latin hypercube design}, as follows: 
	To generate $\nart$ instances, partition the value range of each attribute into $\nart$ equally sized intervals. This yields a grid with $(\nart)^\ndim$ cells. 
	Assign the integers $1, \dots, \nart$ to cells so that each integer appears only once in any dimension of the grid. 
	Now, randomly select an integer $i \in 1, \dots, \nart$. 
	Finally, generate $\nart$ instances by sampling uniformly within the $\nart$ cells which integer $i$  has been assigned to.
\end{definition}
The only generation parameter of $\lhs$ is $\nart$. \cref{fig:examp_lhs} features an illustration of the $\lhs$ approach. 

In our experiments, we let the artificial instances generated with $\lhs$ compete against the output of approaches specifically designed for the generation of outliers. 
In the two use cases \emph{casting task} and \emph{one-class tuning}, we find that the instances generated with $\lhs$ yield comparable outlier-detection quality. 

\subsection{Adversarial Machine Learning}

The recent development of Generative Adversarial Networks (GANs) \cite{Goodfellow2014-vw} has given adversarial machine learning much attention. 
In general, the field is concerned with the robustness of machine learning with respect to adversarial input and countermeasures \cite{Biggio2017-oi}. 
Such adversarial input is artificial data deemed either evasive or poisonous \cite{Kumar2017-uj}. 
Evasive instances fool a trained classifier, yielding the wrong classification (e.g., spam email that is not classified as such). 
Poisonous instances, on the other hand, prevent a classifier from being trained correctly.

In our view, the generation of adversarial input is similar to that of artificial outliers. 
In essence, an outlier which is wrongly classified as normal can be a very useful artificial outlier, as illustrated later in \cref{sec:problem}. 
The fact that there is an outlier-generation approach using GANs~\cite{lee2018-tr,dai2017good} (see \cref{seq:ganGen} for details) further emphasises the strong connection between artificial outliers and adversarial machine learning. 
The idea \citet{Goodfellow2014-vw} introduce as GAN is to have two models, a generative and a discriminative one, that compete against each other. 
The generating model tries to generate instances which the discriminator model cannot tell apart from genuine ones. 
The generative model is thus encouraged to generate instances as close as possible to genuine ones. 
This idea is similar to a generation approach proposed by \citet{Hempstalk2008-tu} (\cref{def:densAprox}). 
However, approaches to generate adversarial inputs tend to be very specific to a classifier or task they are supposed to attack \cite{Brendel2017-vi}. 
Thus, one cannot always use them for the generation of artificial outliers.

\section{Problem Definition}
\label{sec:problem}

One of the central building blocks for a general perspective on artificial outliers is a unified problem formulation. 
Such a formulation that takes into account the different use cases for artificial outliers described in \cref{sec:use_of_arts} now follows.
The integration of artificial outliers within other research fields given in \cref{sec:con_to_other} is important here as well, since it frames the distinctive ideas from this field.

Artificial outliers are expected to approximate instances from $\outdistgenu{}$. 
When we know $\outdistgenu{}$, obtaining artificial outliers becomes trivial: 
We just sample from the distribution that matches our knowledge. 
An exemplary scenario is when we want to detect faults in a system, and the maintainer knows how these faults are distributed. 
However, one usually does not have any or has only very limited knowledge of $\outdistgenu{}$. 
To illustrate, it is highly unlikely in the exemplary scenario just sketched that the distribution of faults is well known without having some faulty instances. 
Thus, we have to rely on assumptions on outliers that allow the generation of instances approximating ones from $\outdistgenu{}$. 
To reflect our limited or missing knowledge on $\outdistgenu{}$, we make assumptions so that outliers generated are as \emph{uninformative} as possible \cite{Theiler2003-dy}. 
That is, they should disclose only very few characteristics of outliers and hence result in the detection of many possible types. 
However, at the same time, we want to make the generated artificial outliers as \emph{interesting} as possible. \cref{def:interesting_outs} formalizes the concept of interesting artificial outliers, and \cref{exp:interesting_place} illustrates it. 

\begin{definition}
	\label{def:interesting_outs}
	A set of \emph{interesting artificial outliers} $\inrset = \{ \inst[1], \dots, \inst[\nart] \}$ is a set of instances that solve a use case for artificial outliers well. 
\end{definition}
Recall the use cases introduced earlier, \emph{casting task}, \emph{one-class tuning} or \emph{exploratory usage}, and consider the following example.

\begin{figure}[ht]
	\centering
	\IfFileExists{./figures/example_interesting.pdf}{
		\includegraphics[width=0.65\linewidth]{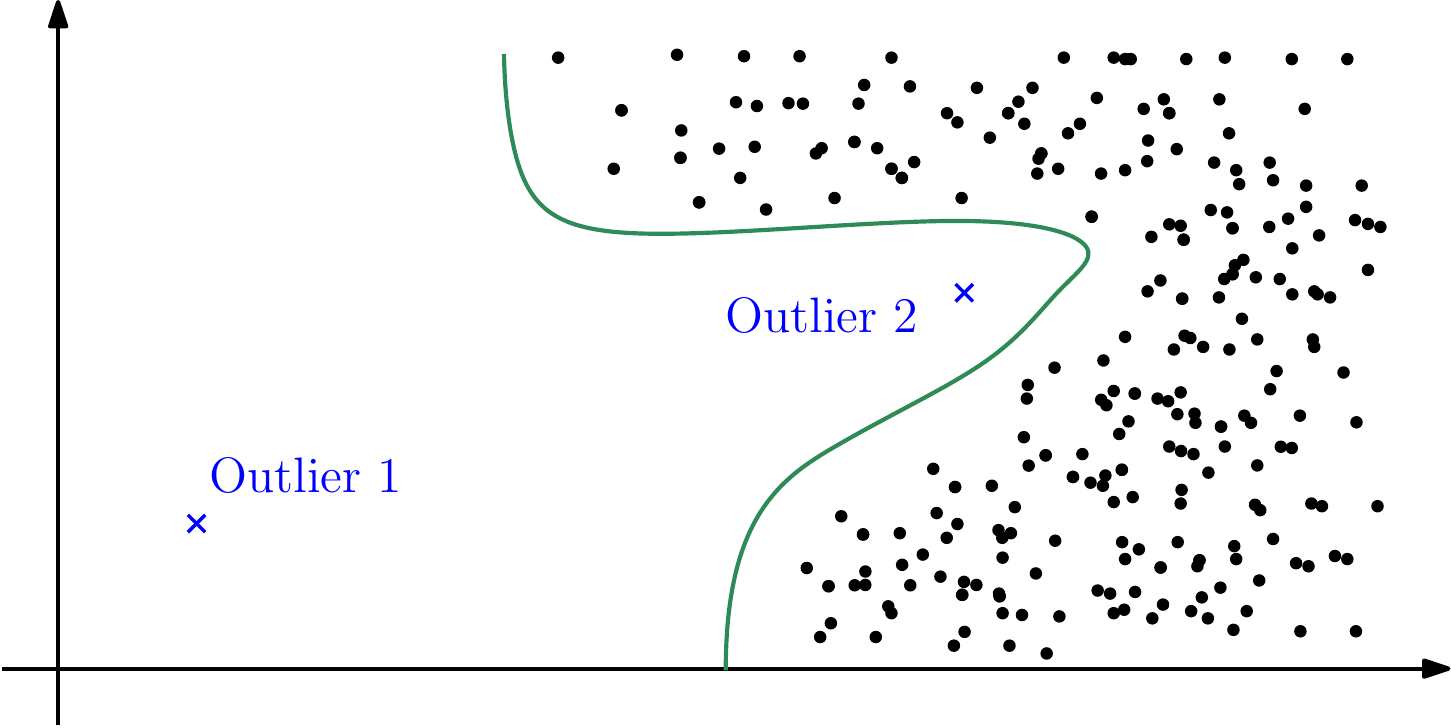}
	}{
		\framebox{\begin{minipage}[c][5cm][c]{1\linewidth} %
				\centering
				{\LARGE \color{red} Image is missing!}
		\end{minipage}}
	}
	\caption{Illustration of interesting artificial outliers.}
	\Description{An illustration of interesting outliers. It shows some normal instances that are separated from the empty instance space by a valley shaped border. Within this Valley is Outlier 2. Far in the empty space is Outlier 1.}
	\label{fig:exp_interesting}
\end{figure}

\begin{example}
	\label{exp:interesting_place}
	The use case in this example is \emph{casting task} (i.e., training a classifier for outlier detection with training data that contains only normal instances). 
	The use case is solved well if the outlier-detection accuracy is later high. 
	To train the classifier, we use  artificial outliers. 
	See \cref{fig:exp_interesting}. 
	The green line is the best decision boundary between normal and outlier instances. 
	Outlier 1 is far from any normal instance. 
	Such an outlier is not very useful when training the classifier. 
	It is rather trivial to classify it as outlying, and it might even pull the decision boundary of the classifier away from normal instances. 
	Outlier 2, by contrast, is helpful when learning the correct decision boundary and is thus rather \emph{interesting}.
\end{example}

The interestingness of artificial outliers depends heavily on the specific application \cite{Steinwart2005-vz,Hastie2009-mu}. 
If we are interested in the exploratory use case instead of the one from \cref{exp:interesting_place}, artificial outliers far away from normal instances might be interesting as well. 
This relationship makes a precise and at the same time general definition of interesting artificial outliers difficult. 
Another issue is that interestingness of artificial outliers also depends on the other generated instances. 
It might well be that Outliers 1 and 2 in combination lead to a better decision boundary. 
\cref{def:interesting_outs} has reflected this possibility.

The situation is even more complex, however, since the number of generated outliers is important as well. 
Any additional artificial instance increases the computational effort. 
Thus, we want to generate as few artificial outliers as possible. 
This leads to the following definition. 

\begin{definition}
	\label{def:optim_int_out}
	A \emph{minimal} set $\inrset$ of artificial outliers is a set of interesting ones that has a minimal number of elements $\nart$ and is still interesting.
\end{definition}
When generating artificial outliers for a use case, one would like to have a minimal set $\inrset$. 
However, there is a trade-off. 
Interesting outliers are often counter to uninformative outliers. 
Consider \cref{exp:interesting_place}, where we suppose that outliers occur close to genuine instances and not everywhere. With such additional assumptions, one clearly loses some generality.
One could argue that having some uninteresting artificial outliers is better than losing this generality. 
However, in high-dimensional spaces in particular, including uninteresting artificial outliers can soon become very expensive computationally \cite{Tax2001-na,Hempstalk2008-tu,Steinbuss2017-hv,Davenport2006-ck}. 
Hence, existing approaches make different assumptions about outliers in order to obtain a minimal set $\inrset$. 
This will become apparent in \cref{sec:place_approach} when we describe the approaches. 
However, having a specific use case in mind, one must be careful that the assumptions actually fit the use case. 
For instance, as mentioned before, if one wants to perform an explorative analysis, generating instances only very close to the boundary of normal instances tends not to be good. 

If artificial outliers are used, usually not only $\outdistgenu{}$ is missing, but also $\normdistgenu{}$. 
Otherwise, a generative model might be preferable, see \cref{sec:gen_models}. 
Hence, the generation is based only on samples from $\normdistgenu{}$ possibly mixed with some from $\outdistgenu{}$, i.e., on $\dataset$. 
Of course, it is possible that the instances from $\dataset$ are not sufficient to represent $\normdistgenu{}$. 
Think of the case that there is no instance from $\dataset$ in a large part of the instance space that should be regarded as normal. 
An artificial outlier in this part might then be an outlier regarding $\dataset$ but not regarding $\normdistgenu{}$. 
However, the important assumption behind all generation approaches is that there is a sufficient number of genuine instances ($\ngenu$) available.

Next to the actual generation of artificial outliers, approaches also exist to filter existing artificial outliers for interesting ones. 
That is, instead of generating instances at a very specific location (for example very close to the boundary), one generates many artificial outliers with some simple approach and tests which ones are interesting. 
Some of these filtering approaches have been proposed together with a specific approach to generate the outliers. 
However, they might also work well when the generating approach is a different one. 
Thus, in the following two sections, we  first describe the different approaches relevant to actually generate artificial outliers and then filtering approaches.

\section{Generating Approaches}
\label{sec:place_approach}

In this section, we review the various generating approaches and put them into context with the problem formulation from \cref{sec:problem}.
This review is another building block of our general perspective. 
We start by classifying the generation approaches in terms of how they relate to the characteristics of $\dataset$. 
We then describe each approach. 
Approaches with a similar generating procedure are described together in order to reduce redundancy and improve comprehension. 
This description results in two somewhat orthogonal classifications of generation approaches: one based on the characteristic of $\dataset$ and one in terms of similar generation procedures.

\begin{figure}[ht]
	\centering
	\resizebox{0.65\linewidth}{!}{%
	\begin{tikzpicture}[
		basic/.style  = {draw, rectangle},
		root/.style   = {basic, rounded corners=2pt, thin, align=center, fill=gray!30},
		level 1/.style={sibling distance=27mm, align=center, basic, rounded corners=2pt, thin},
		level 2/.style={align=center, basic},
		edge from parent/.style={->,draw},
		>=latex]
		
		\node[root] {Model Dependency?}
		child {node[level 1] (c1) {No}}
		child {node[level 1] (c2) {Partly}}
		child {node[level 1] (c3u) {Yes}};
		
		\node[root] [below=1.4cm of c3u] (c3) {Match $\dataset$?}
			child {node[level 1] (c31) {Inverse}}
			child {node[level 1] (c32) {Boundary}}
			child {node[level 1] (c33) {Itself}};
		
		\draw[->,>=latex] (c3u) -- (c3);
		
		\begin{scope}[every node/.style={level 2}]
		\node [below= 0.3cm of c1] (c101) {\unifBox \\ \unifSphere \\ \marginSample \\ \gaussTail};
		
		\node [below of = c2] (c201) {\distBased \\ \boundVal};
		
		\node [below= 0.28cm of c31] (c311) {\invHist \\ \negSelect \\ \infeasExam};
		
		\node [below=0.28cm of c32] (c321) {\boundPlace \\ \negShift \\ \ganGen};
		
		\node [below= 0.28cm of c33] (c331) {\densAprox \\ \surReg \\ \skewBased \\ \maniSamp};
		\end{scope}
		
		\draw[>=latex] (c1) -- (c101);
		\draw[>=latex] (c2) -- (c201);
		\draw[>=latex] (c31) -- (c311);
		\draw[>=latex] (c32) -- (c321);
		\draw[>=latex] (c33) -- (c331);
		
		
		
	\end{tikzpicture}	
}
	\caption{Classification of generating approaches}
	\Description{A tree that classifies the approaches into two five groups. The first group (unifbox, unifSphere marginSample and gaussTail) do not model the data dependency, The second group (distBased and boundVal) partly model the data dependency. Approaches in the remaining three groups all model the data dependency but match the data on different levels. Approaches in one group (invHist, negSelect and infeasExam) match the data inversely, in another group (boundPlace, negShift and ganGen) approaches match the boundary of the data and in the last group (densAprox, surReg, skewBased and maniSamp) approaches match the data itself.}
	\label{fig:class_tree}
\end{figure}
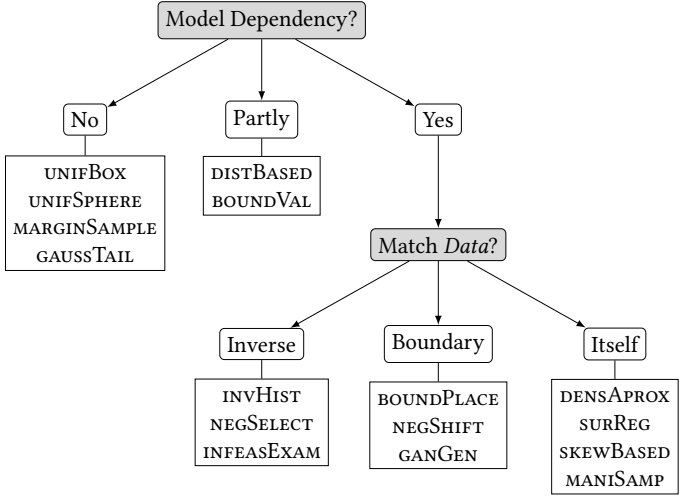

In terms of retaining characteristics from $\dataset$, we group the approaches in six groups, see \cref{fig:class_tree}. 
They differ in the extent of modeling the dependency on $\dataset$, and how well they match with instances from $\dataset$. 
In terms of the dependency, they either do not model it, do so only partly, or model all of it. 
Regarding the match with instances from $\dataset$, the artificial outliers can be somewhat inversely distributed, close to their boundary or entirely similar. 
In the following, we describe the existing approaches, grouped by generation paradigms.

\subsection{Sampling from a Distribution}
\label{sec:sample}

Sampling from a distribution is a common way to generate data. There also exist  approaches to generate outliers with such sampling. The difference among these approaches is the distribution $\outdist$ they sample from.

\subsubsection{Uniform within a Hyper-Rectangle}
\label{sec:unifBox}

\cref{def:unifBox} features the most frequently used distribution \cite{Steinbuss2017-hv,Hastie2009-mu,Hempstalk2008-tu,Shi2006-jh,Abe2006-ca,Steinwart2005-vz,Fan2004-oq,Theiler2003-dy,Fan2001-conf,Tax2001-na,El-Yaniv2007-dv,Davenport2006-ck}.

\begin{definition}[$\unifBox$]
	\label{def:unifBox}
	$\outdistgen{\unifBox}$ is a uniform distribution within a hyper-rectangle encapsulating all genuine instances. 
	The parameters are $\nart$ and the bounds $a, b \in \R^\ndim$ for the hyper-rectangle. 
\end{definition}
Instances from $\dataset$ usually determine the bounds $a, b \in \R^\ndim$. 
For this reason, this approach needs them as input. 
\citet{Tax2001-na} and \citet{Fan2004-oq} state only that these bounds should be chosen so that the hyper-rectangle encapsulates all genuine instances. 
\cite{Steinbuss2017-hv} uses the minimum and maximum for each attribute obtained from $\dataset$. \citet{Theiler2003-dy} mention that the boundary does not need to be far beyond these boundaries. 
\citet{Abe2006-ca} propose the rule that the boundary should expand the minimum and maximum by 10\%. 
\citet{Desir2013-xp} propose to expand the boundary by 20\%. 
In \cref{sec:gen_param}, we describe the boundaries used in our experiments.

\subsubsection{Uniform within a Hyper-Sphere}

\citet{Tax2001-na} propose a straightforward adaptation of the distribution from the $\unifBox$ approach that emphasises generating outliers close to genuine instances.

\begin{definition}[$\unifSphere$]
	\label{def:unifSphere}
	$\outdistgen{\unifSphere}$ is a uniform distribution in the minimal bounding sphere encapsulating all genuine instances. 
	The only generation parameter is $\nart$.
\end{definition}
There are various approaches to obtain or approximate the minimal bounding sphere (e.g., see \cite{Larsson2008-bk}). \citet{Tax2001-na} propose to use the optimization approach also used when fitting a SVDD. 
Sampling uniformly from a hyper-sphere is not simple. 
\citet{Tax2001-na} therefore propose a method using transformed samples from a multivariate Gaussian distribution.

\subsubsection{Manifold Sampling}

\citet{Davenport2006-ck} propose a generation approach that also uses hyper-spheres. 
Similar to $\unifSphere$, the aim is to generate instances close to genuine ones. 
More specifically, they want to generate instances within the manifold in which the genuine instances lie. 
To model this manifold, they use multiple hyper-spheres. 
The sampling distribution $\outdist$ is formalized in \cref{def:maniSamp}.

\begin{definition}[$\maniSamp$]
	\label{def:maniSamp}
	For each $\inst[i] \in \dataset$, let $\textit{avgDist}_i^k$ be the average distance to its $k$ nearest neighbors. 
	Then the sampling distribution $\outdistgen{\maniSamp}$ is the union of the hyper-spheres with center $\inst[i]$ and radius $\textit{avgDist}_i^k$ for $i \in 1, \dots, \ngenu$.
	The parameters are $\nart$ and $k$. 
\end{definition}

\subsubsection{Using Density Estimation}

\citet{Hempstalk2008-tu} try to reformulate the \emph{casting task} use case so that the set of artificial outliers is close to minimal.
For this objective, they find that the ideal distribution of artificial outliers should be the one of the normal instances. 
However, since the distribution of normal instances is usually not known, they propose to estimate it with any density-estimation technique (see \cref{def:densAprox}).

\begin{definition}[$\densAprox$]
	\label{def:densAprox}
	$\outdistgen{\densAprox}$ is the result of density estimation on genuine instances. 
	The parameters are $\nart$ and the density-estimation technique.
\end{definition}
\citet{Hempstalk2008-tu} state that any density-estimation technique can be used in principle, as long as it is possible to draw samples from the density estimate. 
In their experiments, they use two variants. 
In both cases, a certain distribution is assumed and its parameters are estimated from $\dataset$. 
These distributions are as follows:
\begin{itemize}
	\item A multivariate Gaussian distribution having a covariance matrix with only diagonal elements. 
	That is, attributes are independent.\footnote{This actually results in the same generation process Abe et al.~\cite{Abe2006-ca} proposes for the $\marginSample$ approach.}
	\item A product of $\ndim$ Gaussian mixtures, one for each attribute.
\end{itemize}

\subsubsection{Outside of a Confidence Interval}

\citet{Pham2014-lk} find that outlier instances should be very different from normal instances. 
They propose to generate outliers far from most genuine instances by using the distribution from \cref{def:gaussTail}.

\newcommand{\distribution}{\mathcal{D}}
\begin{figure}[ht]
	\centering
	\includegraphics[width=0.75\linewidth]{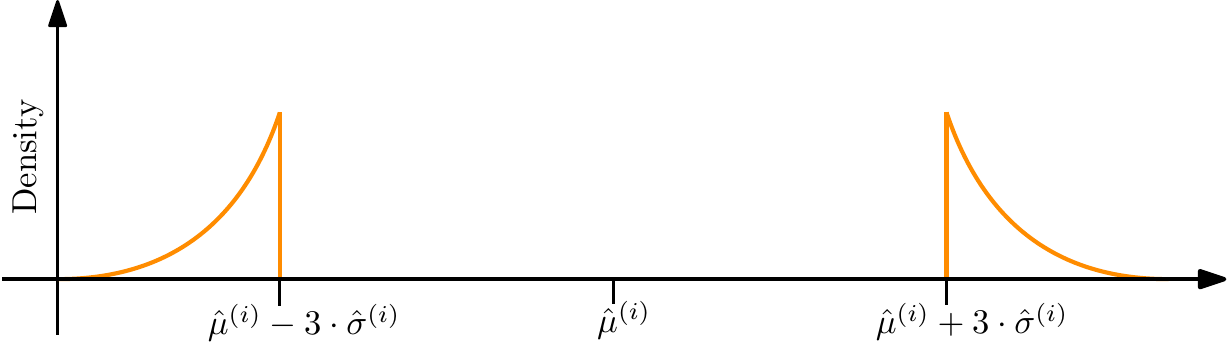}
	\caption{Illustration of $\distribution^{(i)}$ from $\gaussTail$ approach.}
	\Description{Illustration of the distribution gaussTail samples from. Basically, the density of a Gaussian in which the middle part is set to zero.}
	\label{fig:expgausstail}
\end{figure}

\begin{definition}[$\gaussTail$]
	\label{def:gaussTail}
	Let $\hat{\mu}^{(i)}$ and $\hat{\sigma}^{(i)}$ be the mean and standard deviation estimated from Attribute $i$ in $\dataset$. 
	The distribution $\distribution^{(i)}$ has density zero for any value $x \in \hat{\mu}^{(i)} \pm 3 \cdot \hat{\sigma}^{(i)}$. Then $\outdistgen{\gaussTail}$ is the product of $\distribution^{(i)}$ for all attributes. 
	The only parameter is $\nart$.
\end{definition}
\citet{Pham2014-lk}  do not discuss what the density outside the interval $\hat{\mu}^{(i)} \pm 3 \cdot \hat{\sigma}^{(i)}$ should be like.
In our experiments, we assume $\distribution^{(i)}$ to be a Gaussian with the density $x \in \hat{\mu}^{(i)} \pm 3 \cdot \hat{\sigma}^{(i)}$ set to zero. 
See \cref{fig:expgausstail}.

\subsubsection{Inverse Histogram}

\citet{Desir2013-xp} propose the distribution of outliers to be exactly complementary to the distribution of normal instances. 
In other words, they propose to use the distribution from \cref{def:invHist}.

\begin{definition}[$\invHist$]
	\label{def:invHist}
	Let $H_\textit{normal}$ be the normalized histogram of normal instances. Then $\outdistgen{\invHist}$ has pdf $1-H_\textit{normal}$. The parameters are $\nart$ and the histogram-estimation technique.
\end{definition}
\citet{Desir2013-xp} do not discuss details on how to compute the normalized histogram. They do say that the instance-space boundary (i.e., minimum and maximum of each attribute) should be increased by 20\%.

\subsubsection{Generative Adversarial Networks}
\label{seq:ganGen}

\citet{dai2017good} and \citet{lee2018-tr} propose to use a GAN \cite{Goodfellow2014-vw} to generate artificial outliers. 
The generator from a trained GAN architecture is an implicit generative model \cite{dai2017good}. 
Hence, it can generate instances that are similar to the instances it was trained with (the genuine ones) but does not provide a closed form of their density.
The aim of the generator is to maximize the similarity of generated and genuine instances. 
Hence, using the generator to generate outliers is not straightforward. 
To achieve the generation of outliers, \citet{dai2017good} and \citet{lee2018-tr} follow the same strategy. 
A penalty term is added to the objective function of the generator. 
This penalty encourages a generation of instances further away from genuine ones. 
Since both formulations are based on the same idea \cite{lee2018-tr}, \cref{def:ganGen} features the formulation from \cite{dai2017good}. 
We deem it more illustrative for our purpose.

\begin{definition}[$\ganGen$]
	\label{def:ganGen}
	Let $Z$ be the distribution of the prior input noise for the generator function $G\colon \text{supp}(Z)\to \R$. 
	Let $D\colon \R\to [0,1]$ be the discriminator function outputting the probability that an instance is not generated and $p(\cdot)$ an estimate of the density function of genuine instances. 
	$\outdistgen{\ganGen}$ is then the distribution of instances sampled from $G(Z)$ optimized according to
	\begin{equation}
	\label{eq:gan}
	\min_G \ \Ex{\artinst \sim G(Z)}{ \log(p(\artinst))  \ \indicator{p(\artinst) > \varepsilon}}  + \underbrace{\Ex{\artinst \sim G(Z)}{\log(1-D(\artinst))}}_{\text{Original GAN \cite{Goodfellow2014-vw}}}.
	\end{equation}
	The parameters are $\nart$, $\varepsilon$, the network structure for the GAN architecture, and the density estimation technique from which $p(\cdot)$ resulted.
\end{definition}
The first term in \cref{eq:gan} intuitively punishes the generator for generating instances that have a very high density ($> \varepsilon$) according to $p(\cdot)$.
Hence, instances are generated in regions of the instance space with rather low density.

\subsection{Shifting Genuine Instances}
\label{sec:shift}

Approaches in this category modify  attribute values of genuine instances to generate outliers. 
The approaches $\infeasExam$, $\skewBased$ and $\surReg$ use random noise that is added to genuine instances. 
Both $\boundPlace$ and $\negShift$ shift genuine instances so that they move away from other genuine ones.

\subsubsection{Plain Gaussian Noise}

\citet{Neugebauer2016-od} propose to alter instances with Gaussian noise and then filter the resulting instances for those far from normal ones (i.e., having a certain distance to them). 
Only in the first iteration are normal instances altered; then, only the resulting artificial outliers are. 
See $\infeasExam$ in \cref{alg:infeasExam}. 
The approach requires not just genuine instances but genuine normal ones.

\begin{algorithm}
	\caption{$\infeasExam$}
	\label{alg:infeasExam}
	\begin{algorithmic}[1]
		\Parameters $\nart, \ \mu, \ \sigma, \ \alpha, \ \epsilon$
		\For{$i \in 1,\dots,\ngenu$}
		\State $\artinst[i] = \inst[i] + \mathcal{N}(\mu, \sigma) \cdot \alpha$
		\State $\textit{dist}_i = $ distance of $\artinst[i]$ to closest normal instance
		\If {$\textit{dist}_i \geq \epsilon$}
		\State Add $\artinst[i]$ to artificial outliers $\artoutset$
		\EndIf
		\EndFor
		\Repeat 
		\State Randomly choose $\artout[i]$ from $\artoutset$
		\State $\artinst[i] = \artout[i] + \mathcal{N}(\mu, \sigma) \cdot \alpha$
		\State $\textit{dist}_i = $ distance of $\artinst[i]$ to closest normal instance
		\If {$\textit{dist}_i \geq \epsilon$}
		\State Add $\artinst[i]$ to artificial outliers $\artoutset$
		\EndIf
		\Until{$|\artoutset| = \nart$}
	\end{algorithmic}
\end{algorithm}

\subsubsection{Scaled Gaussian Noise}

\citet{Deng2007-rx} propose a so-called skewness-based generation approach for artificial outliers that uses noise added to genuine instances. 
Similar to the case in \cref{alg:infeasExam}, this noise is Gaussian; it is scaled by a parameter~$\alpha$. However, the approach of \citet{Deng2007-rx} does not make use of any filtering or of several iterations. 
See \cref{def:skewBased}.

\begin{definition}[$\skewBased$]
	\label{def:skewBased}
	Let $\hat{\sigma}^{(i)}$ be the standard deviation estimated for Attribute $i \in 1, \dots, \ndim$, and let each $\randval[i]$ be a random value drawn from $\mathcal{N}(0, 1)$. Further, let
	\begin{equation}
	v^{(i)} := \frac{\hat{\sigma}^{(i)}}{\sum_{j=1}^{\ndim} \hat{\sigma}^{(j)}}, \quad
	\textit{noise}^{(i)} := \frac{\randval[i]}{\sum_{j=1}^{\ndim} \randval[j]}.
	\end{equation}
	An outlier $\artout_{\skewBased}$ is then generated by
	\begin{equation}
	\artout_{\skewBased} = \inst + \alpha \cdot (v^{(1)} \cdot \textit{noise}^{(1)}, \dots, v^{(\ndim)} \cdot \textit{noise}^{(\ndim)}),
	\end{equation}
	where $\inst$ is a randomly drawn genuine instance. The parameters are $\nart$ and $\alpha$.
\end{definition}

\subsubsection{Uniform Noise}

\begin{figure}[ht]
	\centering
	\IfFileExists{./figures/exp_surReg.pdf}{
		\includegraphics[width=.65\linewidth]{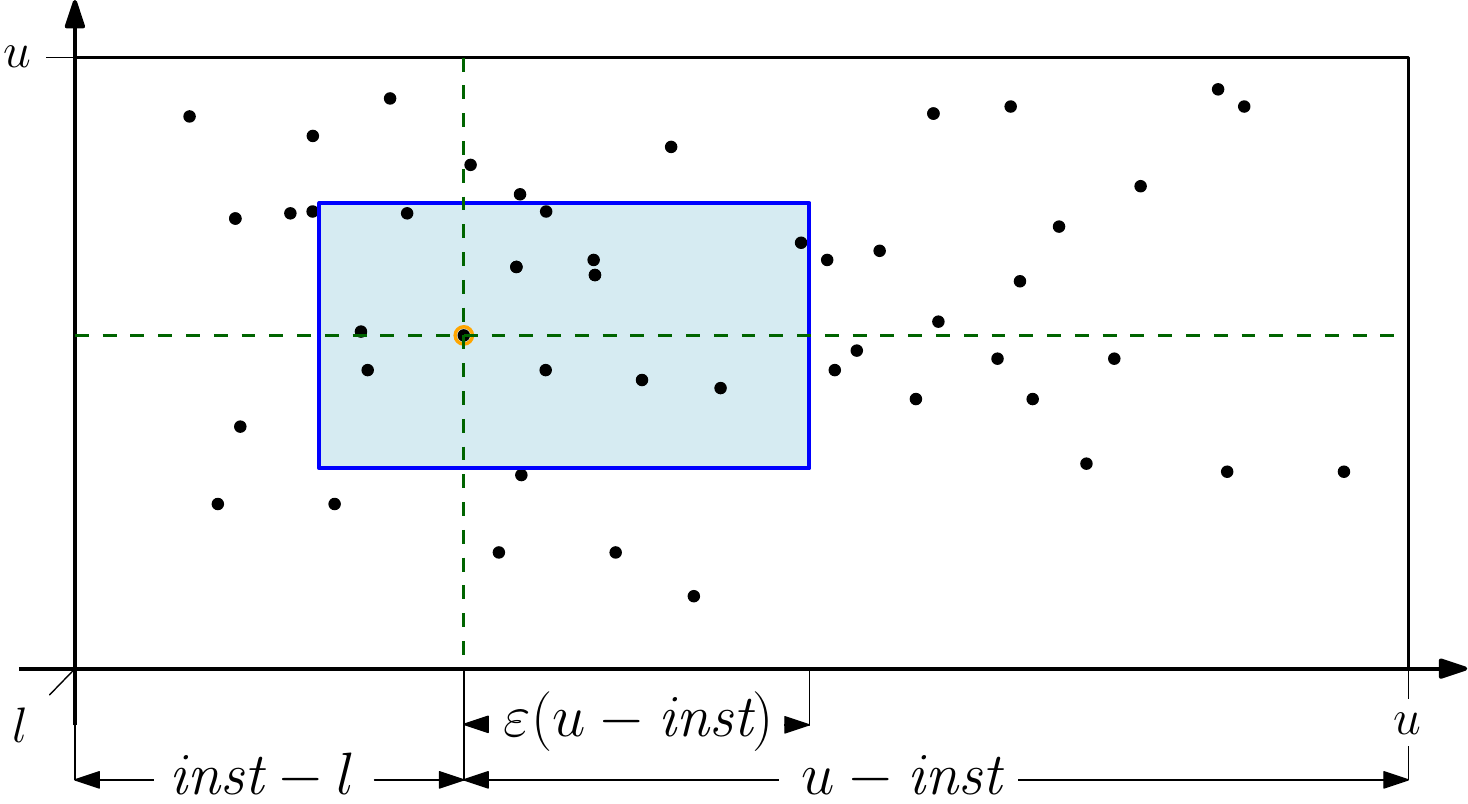}
	}{
		\framebox{\begin{minipage}[c][5cm][c]{1\linewidth} %
				\centering
				{\LARGE \color{red} Image is missing!}
		\end{minipage}}
	}
	\caption{Illustration of the $\surReg$ approach.}
	\Description{Illustration of the surReg approach. Basically, it shows how the uniform distribution surrounding an exemplary genuine instance.}
	\label{fig:surReg}
\end{figure}

\cite{Steinbuss2017-hv} proposes another approach to generate outliers. 
The rationale is to adjust the tightness of artificial outliers around normal instances. 
The approach adds uniform noise parameterized with $\varepsilon \in \left[0, 1\right]$ to genuine instances. 
If $\varepsilon = 1$, the generation will result in samples from a uniform distribution. 
If $\varepsilon = 0$, there will be samples of genuine instances. 
See \cref{fig:surReg} and \cref{def:surReg}.

\begin{definition}[$\surReg$]
	\label{def:surReg}
	Let $\inst$ be a randomly drawn instance from $\dataset$. W.l.o.g., $\inst \in [u,l]^\ndim$. Further, let $\randval[i] \ i \in 1, \dots, \ndim$ be random values drawn uniformly from the range $\varepsilon \cdot (\instval[i] - l)$ to $\varepsilon \cdot (u - \instval[i])$. Then an outlier $\artout_\surReg$ is generated by
	\begin{equation}
	\artout_\surReg = \inst + (\randval[1], \ \dots, \ \randval[\ndim]).
	\end{equation}
	This procedure is repeated until $\nart$ outliers are generated. The parameters are $\nart$ and $\varepsilon$.
\end{definition}
Experiments in \cite{Steinbuss2017-hv} indicate that $0.1$ can be a good value for $\varepsilon$, in particular if there are many attributes. 
Note that for the $\surReg$ approach, the data set must have been normalized to $[0,1]$. 

\subsubsection{Using Boundary Instances}

\begin{figure}[ht]
	\centering
	\IfFileExists{./figures/exp_boundPlace.pdf}{
		\includegraphics[width=0.5\linewidth]{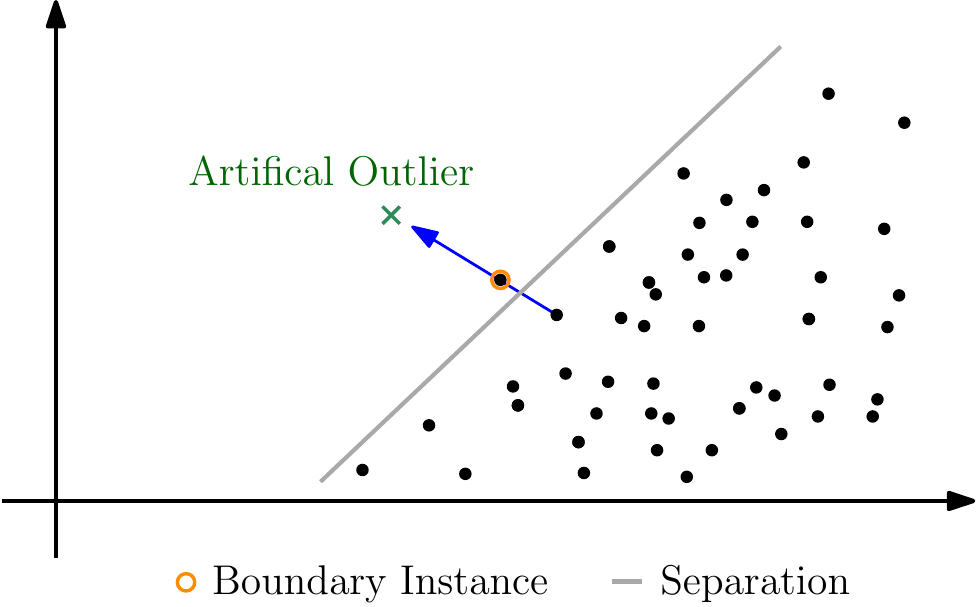}
	}{
		\framebox{\begin{minipage}[c][5cm][c]{0.7\linewidth} %
				\centering
				{\LARGE \color{red} Image is missing!}
		\end{minipage}}
	}
	\caption{Illustration of $\boundPlace$ approach by B\'{a}nhalmi et al.~\cite{Banhalmi2007-pa}.}
	\Description{Illustration of the boundPlace approach. It shows a pile of genuine instances in the lower right that are separated from a boundary instance by a straight line. An arrow indicates the shift of one instance close to this boundary instance such that an artificial outlier is generated.}
	\label{fig:boundPlace_negShift}
\end{figure}

\citet{Banhalmi2007-pa} and \citet{Wang2018-vt} both present a similar idea to generate artificial outliers very tightly around the boundary of genuine instances. 
The idea is to have a two-stage process. 
In the first stage, one finds boundary instances (i.e., instances that \enquote{surround} all other genuine instances). They are then used in the second stage to shift genuine instances away from others.
 See \cref{fig:boundPlace_negShift} for an illustration of the approach of \citet{Banhalmi2007-pa}. 
 The approaches by \citet{Banhalmi2007-pa} and~\citet{Wang2018-vt} differ in the following respects:

\begin{enumerate}
	\item How boundary instances are found. \label{it:bound_detect}
	\item Which instances are shifted. \label{it:what_inst}
	\item The magnitude and direction of the shift. \label{it:inst_shift}
\end{enumerate}

\noindent \citet{Banhalmi2007-pa} propose to determine boundary instances with \cref{alg:boundPlace}. 
The idea is that an instance is on the boundary if it is linearly separable from its $k$-nearest neighbors. 
A hard margin SVM is thus fitted to separate the instance under consideration from its $k$-nearest neighbors. 
If it finds such a separation, the instance is deemed on the boundary.

\begin{algorithm}
	\caption{Boundary detection by B\'{a}nhalmi et al.~\cite{Banhalmi2007-pa}}
	\label{alg:boundPlace}
	\begin{algorithmic}[1]
		\Parameters $k$
		\For{$\inst \in \dataset$}
		\State $\neighset = k$-nearest neighbors of $\inst$ in $\dataset$
		\State $e_i = \frac{\neigh[i] - \inst}{\lVert \neigh[i] - \inst \rVert} \quad \forall \ i \in 1,\dots,k, \ \neigh[i] \in \neighset$
		\State Separate $e_i$ from origin with hard margin SVM
		\If {Separation succeeds}
		\State $\inst$ is boundary instances
		\State Save $\textit{v}_{\inst} = \sum_{i=1}^{k} \alpha_i e_i$ 
		\EndIf
		\EndFor
	\end{algorithmic}
\end{algorithm}

\noindent The vector $\textit{v}_{\inst}$ in \cref{alg:boundPlace} is used to compute the shift direction of genuine instances, see \cref{def:boundPlace}. 
The $\alpha_i$s result from fitting the SVM. 
They weight the contribution of $e_i$ to the final separation.\footnote{See \cite{Banhalmi2007-pa} for two refinements of this approach that increase the maximally possible $\nart$.}

\begin{definition}[$\boundPlace$]
	\label{def:boundPlace}
	Let $\textit{Bounds}$ be the set of boundary instances found with \cref{alg:boundPlace}, with $\textit{V}$ as the set of the vectors $\textit{v}_{\inst}$ saved for each boundary instance found in \cref{alg:boundPlace}. 
	For an instance $\inst \in \dataset \setminus \textit{Bounds}$, let $\textit{bound} \in \textit{Bounds}$ be the closest boundary instance to $\inst$ with $\textit{v}_{\inst} \in \textit{V}$. 
	Let $\Delta := \textit{bound} - \inst$. 
	Further,
	\begin{equation}
	\textit{CosAngle} = \frac{\textit{v}_{\inst}' \cdot (-\Delta)}{\lVert \textit{v}_{\inst} \rVert \cdot \lVert \Delta \rVert }, \quad \textit{Shift} = \frac{\textit{magni}}{\textit{magni} \cdot \textit{curv} + \textit{CosAngle}},
	\end{equation}
	where $\textit{magni}$ and $\textit{curv}$ are parameters. 
	The instance $\inst$ is then shifted by
	\begin{equation}
	\artout_\boundPlace = \inst +  \Delta \cdot \left( 1 + \frac{\textit{Shift}}{\lVert \Delta \rVert} \right).
	\end{equation}
	Then $\artout_\boundPlace$ is an artificial outlier generated with the $\boundPlace$ approach if $\artout_\boundPlace$ is deemed a boundary instance. 
	This procedure is repeated for every $\inst \in \dataset \setminus \textit{Bounds}$. 
	The parameters are $k, \textit{magni}$, and $\textit{curv}$.
\end{definition}

\citet{Wang2018-vt} propose different instantiations for Items~\ref{it:bound_detect}--\ref{it:inst_shift}. 
To detect boundary instances, they rely on \cref{alg:negShift}, the border-edge pattern selection (BEPS) algorithm \cite{Li2011-vg}. 
Like \cref{alg:boundPlace}, BEPS also relies on the $k$-nearest neighbors to decide whether an instance $\inst$ is on the boundary or not. 
However, instead of checking for linear separability using a hard margin SVM, it uses a technical condition on the vectors from a neighbor to the instance $\inst$ ($v_i$). 
See \cite{Li2011-vg,Wang2018-vt} for details.

\begin{algorithm}
	\caption{BEPS Algorithm from Wang et al.\ \cite{Wang2018-vt}}
	\label{alg:negShift}
	\begin{algorithmic}[1]
		\Parameters None
		\State $k = \lceil 5 \, \text{log}_{10} (\ngenu) \rceil, \ \textit{thresh} = 0.1$
		\For{$\inst \in \dataset$}
		\State $\neighset = k$-nearest neighbors of $\inst$ in $\dataset$
		\State $v_i = \frac{\inst - \neigh[i]}{\lVert \inst - \neigh[i] \rVert} \quad \forall \ i \in 1,\dots,k, \ \neigh[i] \in \neighset$
		\State Calculate $\textit{norm} = \sum_{i=1}^{k} v_i$
		\State $\theta_i = v_i' \cdot \textit{norm} \quad \forall \ i \in 1,\dots,k$
		\State $l = \frac{1}{k} \sum_{i=1}^{k} \indicator{\theta_i \geq 0}$
		\If {$l \geq 1 - \textit{thresh}$}
		\State $\inst$ is boundary instances
		\State Save $\textit{norm}$
		\EndIf
		\EndFor
	\end{algorithmic}
\end{algorithm}

\begin{definition}[$\negShift$]
	\label{def:negShift}
	Let $\textit{Bounds}$ be the boundary instances found with \cref{alg:negShift}. Further,
	\begin{equation}
	\textit{scale} = \frac{1}{|\textit{Bounds}| \cdot k} \sum_{\inst \in \textit{Bounds}} \sum_{i=1}^{k} \lvert \inst - \neigh[i] \rvert,
	\end{equation}
	where $\neigh[i]$ is the $i$-th neighbor of an instance $\inst \in \textit{Bounds}$. $\inst$ is then shifted by
	\begin{equation}
	\artout_\negShift = \inst +  \frac{\textit{norm}}{\lvert \textit{norm} \rvert} \cdot \textit{scale},
	\end{equation}
	where $\textit{norm}$ comes from \cref{alg:negShift}. 
	This procedure is repeated for every $\inst \in \textit{Bounds}$. There is no parameter.
\end{definition}
The $\textit{scale}$ value determines how far a boundary instance should be shifted. 
This is based on the distance of each boundary instance to its $k$-nearest neighbors. 
Shifting a boundary instance in the direction of $\textit{norm}$ then generates an artificial outlier.

\subsection{Sampling Instance Values}
\label{sec:samplingInst}

Approaches from this category generate outliers similar to the ones in \cref{sec:shift}. 
They do so by directly using genuine instances. 
However, instead of creating new attribute values, the current approaches recombine existing values of genuine instances to form artificial outliers.

\subsubsection{From the Marginals}

Several articles propose to use marginal sampling to generate outliers~\cite{Shi2006-jh,Theiler2003-dy,Abe2006-ca,Hastie2009-mu}. 

\begin{definition}[$\marginSample$]
	\label{def:marginSample}
	Let $\normdistonenoarg{i}$ be the distribution of Attribute $i$. 
	Then
	\begin{equation}
	\outdistgen[\inst]{\marginSample} = \normdistone{1} \cdot \ \cdots \ \cdot \normdistone{\ndim}.
	\end{equation}
	That is, one can generate outliers from $\outdistgen{\marginSample}$ by sampling a value from each attribute independently. The only parameter is $\nart$.
\end{definition}
From \cref{def:marginSample}, it follows that $\outdistgen{\marginSample}$ and $\normdistgenu{}$ have the same marginal distributions. 
However, in $\outdistgen{\marginSample}$, the attributes are mutually independent, while in $\normdistgenu{}$, they are often not. 
\cref{def:marginSample} gives the distribution outliers are generated with explicitly. 
Thus, $\marginSample$ is closely related to the approaches sampling from a distribution (cf.\ \cref{sec:sample}).

\subsubsection{In Sparse Regions}

\citet{Fan2001-conf,Fan2004-oq} introduce the distribution-based generation approach. 
The following summary is based on our understanding of the respective publications which do not come with an open implementation. 
The idea is to generate outliers close to genuine instances while generating more in sparse regions of the instance space. \citet{Fan2004-oq} speculate that "sparse regions are characterized by infrequent values of individual features". 
Based on this speculation, \citet{Fan2001-conf,Fan2004-oq} propose \cref{alg:distBased} for the generation of outliers.

\begin{algorithm}
	\caption{$\distBased$}
	\label{alg:distBased}
	\begin{algorithmic}[1]
		\Parameters Number of runs
		\For{$i \in 1,\dots,\ndim$}
		\State $\textit{Uniq} =$ unique values for attribute $i$
		\State $\textit{freqVal} =$ most frequent value of attribute $i$
		\State $\textit{countFreq} =$ number of instances with $\textit{freqVal}$
		\For{$\textit{val} \in \textit{Uniq}$}
		\State $\textit{countVal} =$ number of instances with $\textit{val}$
		\For{$j \in \textit{countVal},\dots,\textit{countFreq}$}
		\State Choose $\inst \in \dataset$ randomly
		\State Choose $\rand \in \textit{Uniq} \setminus (\textit{val} \cup \instval[i])$ randomly
		\State $\artout = \inst$ with $\rand$ for attribute $i$
		\EndFor
		\EndFor
		\EndFor
	\end{algorithmic}
\end{algorithm}

\noindent In \cref{alg:distBased}, random instances are drawn from $\dataset$ for each possible value of each attribute.
The number of instances sampled is anti-proportional to the frequency of that value. 
Finally, the value of the sampled instances for an attribute is replaced with a random value from that attribute. 
\cref{alg:distBased} can be run several times to generate more outliers.

\subsubsection{Minimal and Maximal Value}

\begin{algorithm}
	\caption{$\boundVal$}
	\label{alg:boundVal}
	\begin{algorithmic}[1]
		\Parameters $\nart$
		\For{$i \in 1, \dots, \nart$}
		\State Choose attribute $j$ and $m$ randomly
		\State $\textit{maxVal}_j, \textit{maxVal}_m$ maximal value of attribute $j$ or $m$
		\State $\textit{minVal}_j, \textit{minVal}_m$ minimal value of attribute $j$ or $m$
		\State $\textit{new}_j =$ randomly choose $\textit{maxVal}_j$ or $\textit{minVal}_j$
		\State $\textit{new}_m =$ randomly choose $\textit{maxVal}_m$ or $\textit{minVal}_m$
		\State Choose $\inst \in \dataset$ randomly
		\State $\artout[i] = \inst$ with $\textit{new}_j$ and $\textit{new}_m$ for attribute $j$ and $m$
		\EndFor
	\end{algorithmic}
\end{algorithm}

\citet{Wang2009-gg} propose an approach which they call the boundary value method.
The idea is to generate artificial outliers so that they surround the genuine instances in each attribute (see \cref{alg:boundVal}). 
For each artificial outlier, the values of two\footnote{If the data set has only two attributes, our implementation replaces the values of only one attribute.} randomly chosen attributes of a randomly chosen genuine instance are replaced with the minimum or maximum of the corresponding attribute. 
Whether a value is replaced by the minimum or maximum of the attribute is also decided by chance. 

\subsection{Real-Valued Negative Selection}
\label{sec:realSelect}

\begin{figure}[ht]
	\centering
	\IfFileExists{./figures/exp_negSelect.pdf}{
		\includegraphics[width=0.5\linewidth]{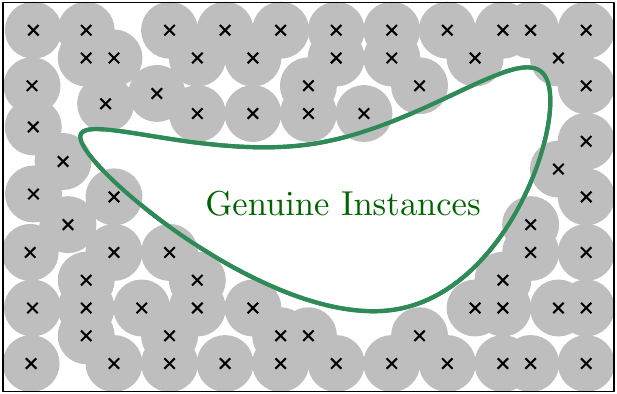}
	}{
		\framebox{\begin{minipage}[c][5cm][c]{0.8\linewidth} %
				\centering
				{\LARGE \color{red} Image is missing!}
		\end{minipage}}
	}
	\caption{Illustration of real-valued negative selection.}
	\Description{The figure displays the banana shaped boundary of genuine instances. This boundary is uniformly surrounded by small circles. A cross marks the centers of the spheres which resembled the artificial outliers.}
	\label{fig:negselect}
\end{figure}

The approach described next does not fit any of the previous categories. 
The generation is based on an adaption of the Negative Selection (NS) algorithm \cite{Forrest1994-bs} from the field of artificial immune systems. 
The idea of NS is inspired by T cells from the human immune system. 
They distinguish cells that belong to the human body (\textit{self}) from ones that do not (\textit{other}). 
In NS, one implements a set of detectors (resembling the T cells) that are then used to distinguish normal (\textit{self}) from outlier (\textit{other}) instances. 
However, NS is usable only when the data set can be represented in binary form, which is to say that each attribute takes the value either 0 or 1.
Thus, \citet{Gonzalez2002-bf} propose the Real-valued Negative Selection (RNS) algorithm (see also \cite{Gonzalez2003-dw}). 
The algorithm tries to find a set of detectors that cover the real-valued instance space not occupied by normal instances. 
Each such detector is a hyper-sphere. 
\cref{fig:negselect} serves as an illustration. 
The green line is the boundary between normal instances and outliers. 
The gray circles are the detectors, with the black crosses as their centers.

\noindent In negative selection, the detectors themselves detect the outlier instances (e.g., by checking whether an instance falls into their vicinity). 
However, \citet{Gonzalez2002-bf} propose the usage of the centres of the detectors as artificial outliers (see \cref{alg:negSelect}). 
An initial set of randomly chosen detectors\footnote{\citet{Gonzalez2002-bf} do not discuss how these are obtained. We simply use the $\unifBox$ approach to this end.} is iteratively optimized. 
In each iteration, the detectors are moved away from genuine instances ($\textit{medDists} < r$) or separated from other detectors ($\textit{medDists} \geq r$) (see \cref{def:negSelect}).

\begin{algorithm}
	\caption{$\negSelect$}
	\label{alg:negSelect}
	\begin{algorithmic}[1]
		\Parameters $\nart, \ r, \ \eta_0, , \ \tau, \ t, \ k, \ \textit{maxIter}, \ \textit{match}(\cdot, \cdot)$
		\State $\textit{Detects} = \nart$ random detectors with age 0
		\For{$\textit{iter} \in 0, \dots, \textit{maxIter}$}
		\State $\eta_\textit{iter} = \eta_0^{-\frac{\textit{iter}}{\tau}}$
		\For{$\textit{detect} \in \textit{Detects}$}
		\State $\neighset = k$-nearest neighbors of $\textit{detect}$ in $\dataset$
		\State $\textit{NeighDists} =$ distances of $\textit{detect}$ to $\neighset$
		\State $\textit{medDists} =$ median of $\textit{NeighDists}$
		\If{$\textit{medDists} < r$}
		\If{age of $\textit{detect} > t$}
		\State Replace $\textit{detect}$ by new random detector
		\Else
		\State Increase age of $\textit{detect}$ by one
		\State $\textit{detect} = \textit{detect} + \eta_\textit{iter} \cdot \textit{dir}_\textit{genu} $
		\EndIf
		\Else
		\State Set age of $\textit{detect} = 0$
		\State $\textit{detect} = \textit{detect} + \eta_\textit{iter} \cdot \textit{dir}_\textit{detect}$
		\EndIf
		\EndFor
		\EndFor
		\
	\end{algorithmic}
\end{algorithm}

\begin{definition}[$\negSelect$] 
	\label{def:negSelect}
	The $\negSelect$ approach is used to generate artificial outliers, outlined in \cref{alg:negSelect}. 
	Here, function $\textit{match}$ is given by
	\begin{equation}
	\textit{match}(\textit{d}_1, \textit{d}_2) = e^{-\frac{\lVert \textit{d}_1 - \textit{d}_2 \rVert^2}{2r^2}}
	\end{equation}
	and $\textit{dir}_\textit{genu}$, $\textit{dir}_\textit{detect}$ by
	{\small \begin{gather}
		\textit{dir}_\textit{genu} = \frac{\sum_{\neigh \in \neighset} \textit{detect} - \neigh }{|\neighset|} \\[3pt]
		\textit{dir}_\textit{detect} = \frac{\sum_{\textit{detect}' \in \textit{Detects}} \textit{match}(\textit{detect}, \textit{detect}') (\textit{detect} - \textit{detect}') }{\sum_{\textit{detect}' \in \textit{Detects}} \textit{match}(\textit{detect}, \textit{detect}')}.
	\end{gather}}
	The parameters are $n_\textit{detects}, \ r, \ \eta_0, , \ \tau, \ t, \ k$, and $\textit{maxIter}$.
\end{definition}
The function $\textit{match}$ in \cref{def:negSelect} determines how well two detectors match (i.e., cover the same instance space), while $\textit{dir}_\textit{genu}$ is the direction in which a detector is shifted to move it away from genuine instances. The direction $\textit{dir}_\textit{detect}$  is used to move a detector away from other detectors.

\subsection{Discussion}

We conclude this section with a summary and a general comparison of the generation approaches presented. 
We have classified the approaches by their connection to the genuine instances (cf.\ \cref{fig:class_tree}) and by the type of procedure used to generate the outliers (\cref{sec:sample,sec:shift,sec:samplingInst,sec:realSelect}). 
The results of our experimental study suggest that for the \emph{one-class tuning} or \emph{casting task} use case, artificial outliers similar to genuine instances (e.g., $\densAprox$ or $\skewBased$) seem to be interesting (cf. \cref{def:interesting_outs}).
Hence, \cref{fig:class_tree} offers a useful resource to guide the selection of a suitable generation approach.

A comparison of approaches within a specific category like \emph{sampling from a distribution} is difficult, since generation approaches tend to differ significantly also within a category. 
For example, within the category just mentioned, the approach based on a simple uniform distribution ($\unifBox$) requires only the attribute bounds to be estimated. 
The approach utilizing GANs ($\ganGen$) in turn requires a deep neural network to be trained. 
Differences like this one arise throughout the surveyed approaches and make finding general benefits or drawbacks difficult. 
Regardless of these difficulties, we give some general results from comparisons in the following.
The category described in \cref{sec:sample} comprises the highest number of approaches. 
Their joint idea is to first fit a distribution to the data
and then generate artificial outliers by sampling from this distribution. 
The fitting of the distribution can be quite resource-intensive, for instance for the $\ganGen$ approach, but it is easy to generate any amount of artificial outliers with sampling. 
As mentioned, not all approaches require many resources to fit the distribution, though.
Approaches that generate outliers by shifting genuine instances (\cref{sec:shift}) or sampling instance values (\cref{sec:samplingInst}) usually require less computational effort in advance of the generation of outliers. 
Additionally, the direct use of genuine instances tends to yield artificial outliers close to these genuine instances (cf.\ \cref{fig:class_tree}).
Sampling instance values  (\cref{sec:samplingInst}) has similar drawbacks and benefits, but is simple to perform.
The $\negSelect$ approach is the only one described in \cref{sec:realSelect}. 
It features a generation paradigm that differs substantially from the procedure of other approaches. 
The iterative optimization of the initial set of artificial outliers is resource-intensive, but does not allow for the straightforward generation of more artificial outliers \emph{a posteriori}, unlike approaches that sample from a distribution. 

In summary, we find it difficult to say which procedure or connection to genuine instances is preferable. 
All approaches presented incorporate ideas that can be useful, or they generate interesting artificial outliers.

\section{Filtering Approaches}
\label{sec:filter}

Having described the existing generation approaches, we now turn to the approaches that filter generated instances for interesting ones.
This is the last building block for our general perspective on artificial outliers.

We group the filter approaches in two groups: those that use a classifier and those that compute and use some statistic. 
\enquote{Artificial instances} refer to instances generated, and \enquote{artificial outliers} to those resulting from filtering the artificial instances. 
Therefore, the artificial outliers should be interesting and close to minimal (see \cref{def:interesting_outs,def:optim_int_out}).

\subsection{Using a Classifier}

\citet{Fan2004-oq} propose an iterative approach to filter artificial instances so that they are further from genuine ones (see \cref{alg:filt_fan}). 
In each iteration, a classifier is trained to distinguish between the genuine instances from $\dataset$ and the generated instances. 
Then, the artificial instances that are classified as genuine are replaced with newly generated instances. 
This process is repeated until only very few artificial instances are removed in an iteration.

\begin{algorithm}
	\caption{Filter using a classifier by Fan et al.~\cite{Fan2004-oq}}
	\label{alg:filt_fan}
	\begin{algorithmic}[1]
		\Parameters $\dataset$, Generation Approach, Classifier Model, $\textit{maxRem}$
		\State $\artinstset = \nart$ artificial instances
		\Repeat 
		\State Train classifier with $\artinstset \cup \dataset$
		\State $\textit{removed} = 0$
		\For{$\artinst \in \artinstset$}
		\State Predict class of $\artinst$ with classifier
		\If {Predicted class is genuine}
		\State Replace $\artinst$ with new generated instance
		\State $\textit{removed} = \textit{removed} + 1$
		\EndIf
		\EndFor
		\Until{$\textit{removed} \leq \textit{maxRem}$}
	\end{algorithmic}
\end{algorithm}

\citet{Abe2006-ca} apply what they call ensemble-based minimum margin active learning. 
It combines the ideas of query by committee and of ensembles. 
Several classifiers are trained one after another, each one on a sample of the genuine and the artificial instances. 
At the end, all classifiers are combined in an ensemble, yielding the final classifier. 
The filter is the sampling procedure that selects the instances used to train a new ensemble member. 

\begin{definition}[Filter with Query by Committee]
	\label{def:filter_abe}
	Let $\classifierset = \{ \classifier_1, \dots, \classifier_m \}$ be a set of $m$ classifiers that have been trained one after another.
	Let $\classifier^\textit{out}(\inst)$ be the probability that Classifier $\classifier \in \classifierset$ classifies $\inst$ as an outlier. Analogously, $\classifier^\textit{norm}(\inst)$ is the probability of $\classifier$ classifying $\inst$ as normal. 
	Let
	\begin{equation}
	\textit{margin}(\classifierset, \inst) = \sum_{\classifier \in \ \classifierset} \classifier^\textit{out}(\inst) - \classifier^\textit{norm}(\inst)
	\end{equation}
	and
	\begin{equation}\label{equ:filter_abe_gauss}
	\textit{gauss}(\mu, \sigma, \xi) = \int_{\xi}^{\infty} \frac{1}{\sigma \sqrt{2\pi}} e^\frac{-(x-\mu)^2}{2\sigma^2} dx,
	\end{equation}
	where \cref{equ:filter_abe_gauss} is used only to simplify \cref{equ:filter_abe_gauss_filled}.
	Then, the \emph{filter with query by committee} is as follows: 
	An instance $\inst \in \datasetext$ is kept with probability
	\begin{equation}\label{equ:filter_abe_gauss_filled}
	\textit{gauss} \left(\mu = \frac{m}{2}, \ \sigma = \frac{\sqrt{m}}{2}, \ \xi = \frac{m + \textit{margin}(\classifierset, \inst)}{2 }\right).
	\end{equation}
\end{definition}
The function $\textit{margin}(\cdot)$ computes the disagreement among the classifiers on the class of $\inst$. 
The function $\textit{gauss}$ transforms this disagreement into a probability and is similar to the CDF of a Gaussian distribution. 
Thus, artificial instances for which there is much disagreement among the classifiers are kept with a higher likelihood. 
Note that the filtering from \cref{def:filter_abe} is probabilistic.
That is, running the filter again can lead to other artificial outliers. 
A very similar idea has been proposed by \citet{Curry2009-bu}. 
However, instead of the filtering from \cref{def:filter_abe}, they make use of the so-called balanced block algorithm \cite{Curry2007-dx}.

\subsection{Using a Statistic}
\label{sec:filter_stat}

Instead of using a classifier to filter artificial instances, several filtering approaches make use of a statistic computed based on the artificial and the genuine instances.  

\cref{def:boundPlace} has featured the filter introduced in \cite{Banhalmi2007-pa}. 
The statistic computed comes from \cref{alg:boundPlace}, which checks whether an instance is a boundary instance. 
Any generated instance that is not a boundary instance according to \cref{alg:boundPlace} is filtered out. 
The filter proposed in \cite{Neugebauer2016-od} has also been described already, in \cref{alg:infeasExam}. 
The statistic computed is the distance from an artificial instance to its nearest genuine neighbor. 
Only if this distance is greater than a certain threshold is the artificial instance deemed an outlier.

\citet{Davenport2006-ck} propose a filter which they call thinning. 
The idea is to filter artificial instances so that the remaining ones are well spread across the whole instance space and have a distance to each other that is as large as possible. 
This proposal resembles the idea behind the $\lhs$ approach from \cref{def:lhs} (see \cref{def:thinning}).

\begin{definition}[Filter through Thinning]
	\label{def:thinning}
	Let $\textit{dist}_{i, j}$ be the Euclidean distance between two artificial instances $\inst[i]$ and $\inst[j]$. 
	Then the \emph{filter through thinning} is as follows:
	\begin{enumerate}
		\item Find the $i \neq j \in 1, \dots, \nart$ for which $\textit{dist}_{i, j}$ is the smallest.
		\item Remove the instance from $\{\inst[i], \inst[j]\}$ which has a lower distance to its nearest neighbor.
	\end{enumerate}
\end{definition}
The filter from \cref{def:thinning} must be applied several times to ensure evenly spread artificial outliers.

\cref{def:filter_stein} is another filtering method, proposed in \cite{Steinbuss2017-hv}. 
The objective has been to filter the artificial instances hidden in certain attribute subsets (i.e., not detectable as outliers in some subsets, but detectable in others). 
Hence, the approach checks for each such attribute subset with some unsupervised outlier-detection technique if the artificial instance is outlying or normal with respect to the genuine instances. 
The unsupervised outlier-detection technique can be thought of as a statistic computed on the instances. 
For example, with the (\emph{k}, \emph{dmax})-outlier definition \cite{Kollios2003-nl}, an instance is an outlier if at most $k$ other instances are closer to it than to $\textit{dmax}$.

\begin{definition}[Filter with Unsupervised Detection]
	\label{def:filter_stein}
	Let $\textit{OutDet}(\cdot)$ be an unsupervised outlier-detection technique; that is, given a set of instances, $\textit{OutDet}(\cdot)$ determines which ones are normal or outlying. 
	The \emph{filter with unsupervised detection} works as follows:
	An artificial instance $\inst$ is kept if either $\textit{OutDet}(\inst) = $ outlier or $\textit{OutDet}(\inst) = $ normal holds. 
	Whether normal or outlier has to hold is a parameter of this filter. 
\end{definition}
Note that in \cite{Steinbuss2017-hv}, depending on the attribute subset, $\textit{OutDet}(\cdot)$ sometimes filters artificial instances deemed outlying and sometimes instances deemed normal. 
To obtain artificial outliers that are rather far away from genuine instances, similarly to \cref{alg:filt_fan}, one only needs to filter for artificial instances $\textit{OutDet}(\cdot)$ deems outlying. 

\subsection{Discussion}

The number of approaches to filter generated instances is much smaller than the number of approaches to generate them. 
We categorize the known approaches in two groups: in one group,  a classifier is used to filter generated instances, and in the other group, certain statistics. 
Filter approaches that use a classifier are usually more time-consuming, since the training of the classifiers has to happen multiple times. 
The filter approaches that utilize statistics are usually much faster. 
However, the filter using unsupervised outlier detection methods can be computationally heavy as well, depending on the detection method used.

Whether it makes sense to use a filter approach ultimately depends on the use case and the approach used to generate artificial outliers. 
The thinning filter (\cref{def:thinning}), for instance, can be quite useful in the \emph{casting task} use case. 
It renders the artificial outliers more uniformly distributed within the instance space, which can be advantageous for that use case \cite{Steinwart2005-vz}.

\section{Experiments}
\label{sec:experiments}

So far, we have presented our general perspective on artificial outliers, including the various approaches to generate artificial outliers in \cref{sec:place_approach} in particular. 
We now experiment with them for insights that also extend to a practical level.
We list three aims behind such experiments.

\emph{Aim 1:} To our knowledge, most presented generation approaches have never been compared to each other systematically.
We aim to make exactly this comparison.

\emph{Aim 2:} \cref{sec:descp_one_class_optim} has explained that the two use cases \emph{casting task} and \emph{one-class tuning} are similar. 
Both result in outlier detection based on classification. 
Thus, another aim of our experiments is to study the quality difference in the resulting detection. 
Not only have  both use cases received much more attention in the literature than the third use case, \emph{exploratory usage}, but we are also unaware of any design of experiments which have taken place to compare artificial outliers in terms of an exploratory usage. 
For this reason, our experiments focus on the \emph{casting task} and \emph{one-class tuning} use cases.
Since both use cases are based on classification, we refer to them by the respective classifiers.

\emph{Aim 3:} Some characteristics of certain types of artificial outliers in terms of the underlying data set are known. 
For example, consider that the outlier-detection quality with some generation approaches decreases with an increasing number of attributes \cite{Tax2001-na,Hempstalk2008-tu,Steinbuss2017-hv,Davenport2006-ck}. 
In our experiments we also analyze these characteristics more closely, for example how prominent such effects are.

In the remainder of this section, we first describe the workflow of our experiments. 
We then describe the data sets and classifiers used and discuss the parametrization of the generation approaches. 
Next, we describe the statistical tools we use to analyze our experimental results. 
We then describe general outcomes from the experiments. 
Finally, we analyze the results in terms of the different classifiers, the generating approaches used, and the data-set characteristics.

\subsection{Workflow}

\cref{alg:experiments} is the workflow for our experiments. 
For each data set and each classifier, we use the generating approaches presented in this survey for training. 
We then test each classifier on all types of outliers.
With \emph{types of outlier} we refer to outliers generated with some approach as well as the genuine outliers. 
For instance, one type of outliers is \enquote{genuine outliers}, while another is \enquote{artificial outliers generated with $\unifBox$}.
The label for training or testing the classifier is whether an instance is a normal genuine instance or an outlier.
To evaluate a detection, we use the Matthews correlation coefficient (mcc), which is particularly suited if the classes can be imbalanced~\cite{Boughorbel2017-ke}. 
This is essentially the correlation between predicted and ground-truth instance labels. 
Each such experiment is repeated 20 times. 
The code for our experiments is publicly available.\footnote{Available at \url{ipd.kit.edu/mitarbeiter/steinbussg/exp-artificial-outliers-FINAL-V3.zip}.}

\begin{algorithm}
	\caption{Experiment Workflow}
	\label{alg:experiments}
	\begin{algorithmic}[1]
		\Require A set of data sets $\textit{Datas}$, a set of classifiers $\classifierset$ and a set of generation approaches\footnotemark $\textit{Generations}$.
		\For{each $\dataset \in \textit{Datas}$}
		\For{each $\classifier \in \classifierset$}
		\State $\textit{Norms} =$ normal instances from $\dataset$
		\State $\textit{TrainNorms} =$ random sample of $\textit{Norms}$ with 70\% of $\textit{Norms}$ size
		\State $\textit{TestNorms} = \dataset \setminus \textit{TrainNorms}$
		\State $\genuoutset =$ genuine outliers from $\dataset$
		\For{each $\textit{gen}(\cdot) \in \textit{Generations}$}
		\State Train $\classifier$ with $\textit{gen}(\textit{TrainNorms}) \cup \textit{TrainNorms}$
		\For{each $\textit{gen}(\cdot) \in \textit{Generations}$}
		\State Predict class of $\textit{gen}(TrainNorms) \cup \textit{TestNorms}$ with $\classifier$
		\State Save Matthews correlation coefficient (mcc)
		\EndFor
		\State Predict class of $\genuoutset \cup \textit{TestNorms}$ with $\classifier$
		\State Save Matthews correlation coefficient (mcc)
		\EndFor
		\EndFor
		\EndFor
	\end{algorithmic}
\end{algorithm}
\footnotetext{A generation approach in this algorithm is represented by a function $\textit{gen}(\cdot)$ which has only a data set as input. Some approaches require additional inputs. See \cref{tab:experiments_overview} for their values.}

\subsection{Data Sets Used}

The data sets we use is a suite of common outlier-detection-benchmark data sets. 
We use most data sets proposed in \cite{Campos2016-zv}. 
These are mostly classification data sets in which one class is deemed outlying. 
We exclude the data sets Arrythmia and InternetAds due to their very high number of attributes (259 and 1555). 
These data sets would extremely increase the runtime of our experiments. 
We add, however, the musk2 data sets from \cite{Dua-2017} with a reasonable number of attributes. 
This addition leaves us with the data sets displayed in \cref{tab:dataset_wilcox}. 
As proposed in \cite{Campos2016-zv}, each data set is scaled so that $\dataset \in [0,1]^\ndim$, and duplicate instances are removed. 
To reduce the run time of our experiments, data sets with more than 1000 instances are downsampled to 1000 instances. 
The outlier and normal classes are downsampled so that their initial ratio remains. 
With $\dataset$ we refer to a data set where each labeled outlier is removed.

\subsection{Classifiers Used}
\label{sec:used_class}

The difference in the use cases \emph{casting task} and \emph{one-class tuning} from \cref{sec:use_of_arts} is that \emph{casting task} uses a binary classifier and \emph{one-class tuning} a one-class classifier. 
There is one family of classifiers that exists in the binary case as well as in the one-class case, namely SVMs. 
For this reason, we use SVMs in our experiments. 
While the one-class SVM only needs a single class of instances (e.g., normal ones) for training, the binary SVM needs two. Both SVMs use linear separation to perform their classification. 
Projecting the data into a kernel space generalizes the linear separation to take any non-linear form (see \cite{Hastie2009-mu}). 
The binary SVM tries to find the best separation of the two classes. 
A common version of the one-class SVM, in turn, tries to separate all available instances from the origin of the transformed space. 
This trick allows the one-class SVM to train with instances from only a single class (see \cite{Scholkopf2001-xx}).

The binary as well as the one-class SVM have two different formulations. 
The binary SVM can be formulated as C-SVM or $\nu$-SVM \cite{Chang2001-oe}. 
The main difference is that they feature different parameters. 
While the C-SVM features a parameter $C \in (0, \infty)$, the $\nu$-SVM features $\nu \in (0,1]$. 
In our experiments we use both, as described later. 
The one-class SVM is formulated as described above in \cite{Scholkopf2001-xx} and is called $\nu$-support classifier. 
\citet{Tax2001-na} formulate another version, the SVDD. 
The two types, however, give identical decision functions when using the Gaussian kernel \cite{Lampert2009-xx}. 
We use the formulation from \cite{Scholkopf2001-xx} in our experiments with this kernel.

\begin{algorithm}
	\caption{Hyperparameter tuning}
	\label{alg:param_select}
	\begin{algorithmic}[1]
		\Parameters $\datasetext, \ \nu_\text{range}, \ s_\text{range}$
		\State Set $\textit{Err}_\textit{Best} = \inf$
		\For{each hyperparameter combination $(\nu, s)$}
		\State Train SVM with $(\nu, s)$
		\State $\textit{Err}_\textit{Art} =$ error on artificial outliers from $\datasetext$.
		\State $\textit{Err}_\textit{Genu} =$ error on genuine instances from $\datasetext$.
		\State $\textit{Err} = 0.5 * \textit{Err}_\textit{Art} + 0.5 * \textit{Err}_\textit{Genu}$
		\If {$\textit{Err}_\textit{best} > \textit{Err}$}
		\State $\textit{Err}_\textit{Best} = \textit{Err}$
		\State $(\nu_\textit{Optimal}, s_\textit{Optimal}) = (\nu, s)$
		\EndIf
		\EndFor
	\end{algorithmic}
\end{algorithm}

The aim with the one-class tuning use case is to find optimal hyperparameters. 
For this search, we use the approach from \cite{Wang2018-vt} displayed in \cref{alg:param_select}. 
It is a grid search over hyperparameters $\nu$ and $s$. 
The values with the lowest error are chosen as the final model. 
As in\ \cite{Wang2018-vt}, we use $\nu_\text{range} = \{0.001, 0.05, 0.1\}$ and $\ s_\text{range} = \{10^{-4}, 10^{-3}, \dots, 10^{4}\}$. 
This approach is referred to as \emph{one-class}. 
For the binary SVM in the \emph{casting task} use case, we have implemented two approaches. One approach is to just use a C-SVM with the default values from the respective implementation $C = 1$ and $s = \frac{1}{\ndim}$, subsequently referred to as \emph{binary}. 
The second approach is to optimize $\nu$ and $s$ of a $\nu$-SVM using \cref{alg:param_select}. 
We refer to this as \emph{binaryGrid}. In summary, we use three types of classifiers: \emph{binary} and \emph{binaryGrid} for the \emph{casting task} use case and \emph{one-class} for the \emph{one-class tuning} use case.

\subsection{Generation Approaches}
\label{sec:gen_param}

\begin{table}[ht]
	\centering
	\caption{Overview of used parameters and approaches. If applicable $\nart$ is set to $\ngenu$.} 
	\label{tab:experiments_overview}
	\begin{tabular}{lrr} 
		\toprule
		Approach & Suggested Parameter Value(s) & Used Parameter Value\\ 
		\midrule 
		$\unifBox$ & Increase of bounds: 0\%, 10\%, 20\% &10\% \\
		$\lhs$ & --- & ---\\
		$\unifSphere$ & --- & --- \\
		$\maniSamp$  & Number of nearest neighbors: 10 & 10\\
		$\marginSample$ &  --- & --- \\
		$\boundVal$ &  --- & --- \\
		$\densAprox$ &  Single Gaussian or mixture& Single Gaussian \\
		$\surReg$ & $\varepsilon = 0.1$  & 0.1\\
		$\skewBased$ & $\alpha = 2$ & 2 \\ 
		$\negShift$ &  --- & --- \\
		$\gaussTail$ &  --- & --- \\
		\bottomrule
	\end{tabular}
\end{table}

We do not vary the parameters of the generation approaches but use their default values if applicable. They are listed in \cref{tab:experiments_overview}. 
The parameter $\nart$ is always set to $\ngenu$ in our experiments. 
That is, the number of genuine and artificial instances is equal if the approach has this parameter. The bounds for the $\unifBox$ approach are extended by 10\%, as proposed in~\cite{Abe2006-ca}. 
We find this a good compromise between no extension and the 20\% increase proposed in~\cite{Desir2013-xp}. 
We perform density estimation in the $\densAprox$ approach with a multivariate Gaussian having a covariance matrix with only diagonal elements. 
For the $\surReg$ approach, we choose $\varepsilon = 0.1$, as suggested by the experiments in \cite{Steinbuss2017-hv}. 
We exclude the $\invHist$ and $\infeasExam$ approaches from our experiments, since they have been proposed specifically for data sets with very few attributes. 
We also have excluded $\boundPlace$, $\distBased$, and $\negSelect$, because of enormous runtimes of our respective implementations, which an experienced programmer from our institution has put together. 
A single execution of $\distBased$ --- the fastest approach of these three excluded ones --- takes more than 50 seconds on a data set with 30 attributes and 650 genuine instances. 
This is roughly the average size of the data sets in our experiments. 
We have to execute each approach for 20 iterations, 16 data sets, and for training as well as for testing three classifiers in combination with 11 other generation approaches (cf.\ \cref{alg:experiments}).
With the $\distBased$ approach, the runtime is very high because our data sets are not categorical. 
Thus, counting the number of occurrences of the values of all attributes in addition to the procedure to look up a new random value becomes very expensive. 
The requirement \enquote{low runtime} also is the reason that we have not implemented nor included $\ganGen$. 
Training a GAN architecture is extremely resource-intensive.

\subsection{Statistical Tools}

A direct comparison of the mcc scores is not very useful due to the many factors influencing the scores. 
Hence, we want to analyse our results statistically by performing an Analysis of Variance (ANOVA) \cite{anovaenz}. 
This analysis allows us to check whether and how strongly the experimental parameters affect the mcc scores. 
We then analyze these effects in more detail with a \emph{post hoc} analysis \cite{McHugh2011-zj}. 
Finally, the Kendall's Tau coefficient \cite{prokhorov2001kendall} is used to determine the effect of certain data characteristics. 

\subsubsection{ANOVA}

From \cref{alg:experiments}, we see that there are four parameters for a specific experiment. 
The classifier ($\classifier$), the underlying data set ($\dataset$), the generation approach the classifier is trained with ($\trainGen$), and the type of outliers the classifier is tested on ($\testGen$). 
We refer to these four as main factors. 
The ANOVA partitions the variation of a dependent variable, here the mcc score, according to so-called sources. 
There are three types of sources: the main factors just mentioned, their interactions, and the residuals. 
The interactions between main factors, denoted by $\textit{Inter}(\cdot)$, are used to account for the joint effect of several main factors, for example if the choice of classifier is not independent of the underlying data set ($\textit{Inter}(1, 4)$ in \cref{tab:anova}). 
To explain residuals, observe that the basis of the ANOVA is regression. 
The residual source is the variation that cannot be accounted for using this regression. 
Each source except for the residual one has a specific number of levels. 
A level of a source is a particular value that it takes. 
For the main factor $\classifier$, for example, the levels are \emph{binary}, \emph{binaryGrid}, and \emph{one-class}.
To perform the ANOVA, one obtains the sum of squares attributed to the different sources from the regression model. 
These and their degrees of freedom are used to compute the F-Value. 
The number of degrees of freedom of a source is given by the number of levels of the source minus one. 
For example, the classifier has three levels. 
Hence, the number of degrees of freedom for this source is 2. 
With the \emph{F}-value, one can perform a statistical test, given in \cref{hyt:anova_f_test}. 
The p-value of this test is computed using the F distribution parameterized by the number of degrees of freedom --- the distribution of the F-Value under the null hypothesis.

\begin{hyptest}[ANOVA \emph{F}-test]
	\label{hyt:anova_f_test}
	Let $\mu_i$ be the mean of the dependent variable for level $i$ from a source with $L$ levels. Then the null and alternative hypothesis of the ANOVA \emph{F}-test are
	\begin{equation}
	\text{H}_0 : \forall \ i, j \in  1, \dots, L: \mu_i = \mu_j
	\end{equation}
	and
	\begin{equation}
	\text{H}_A : \exists \ i, j \in  1, \dots, L: \mu_i \neq \mu_j.
	\end{equation}
\end{hyptest}
The ANOVA \emph{F}-test thus checks whether the mean of at least one level is different from the mean of the other levels. 
In addition to this test, one can compute the partial omega squared ($\omega^2_P$) values \cite{Olejnik2003-tu} using the ANOVA. 
The value for $\omega^2_P$ gives the importance of the respective source in explaining the variation in the mcc score. 
A high $\omega^2_P$ means that this source accounts for a rather large part of the variation in the mcc score.

\subsubsection{Post Hoc Analysis}

Following an ANOVA, one usually performs a post hoc analysis \cite{McHugh2011-zj}. 
Any source for which the \cref{hyt:anova_f_test} is significant is analysed in more detail. 
With the ANOVA \emph{F}-test, one can conclude only that the mean of at least one level differs significantly from the mean of at least one other level. 
However, it usually is interesting for which levels this is the case. 
Hence, for each pair of levels in a significant source, one computes whether there is a difference or not. 
This is what is done in a \emph{post hoc} analysis. 
Clearly, the pairwise tests in a \emph{post hoc} analysis are a case of multiple testing~\cite{McHugh2011-zj}. 
Hence, the \emph{p}-values need to be adjusted accordingly. 
We use the Holm--Bonferroni method \cite{Holm1979-gc} to this end. 
Usually a simple Student's \emph{t}-test is used to compare the means of two levels~\cite{McHugh2011-zj}. 
However, since there are many significant interactions in our ANOVA result, the assumptions behind the Student's \emph{t}-test are usually violated. 
Hence, we use a non-parametric alternative: the Mann--Whitney \emph{U} test \cite{Mann1947-tv}. 	
\begin{hyptest}[Mann--Whitney \emph{U} Test]
	\label{hyt:wilc_test}
	Let $L_i$ and $L_j$ be the distribution function of the dependent variable within Levels $i$ and $j$ of a source. 
	Then the null and alternative hypothesis of the Mann--Whitney \emph{U} test are
	\begin{equation}
	\text{H}_0 : L_i(x) = L_j(x) \ \forall \ x \in [0,1]
	\end{equation}
	and
	\begin{equation}
	\text{H}_A : L_i(x) > L_j(x) \ \text{or} \ L_i(x) < L_j(x) \ \forall \ x \in [0,1].
	\end{equation}
	That is, one variable is stochastically larger or smaller than the other one.
\end{hyptest}
A level that is stochastically greater than another indicates a preference. 
To illustrate, if the one-class classifier is stochastically greater than the binary one, it usually yields higher mcc scores. 
Along with the pairwise test from \cref{hyt:wilc_test}, we provide level-wise means, medians, and density plots of the mcc score when applicable. 
This indicates the direction of the stochastic order. 
For greater clarity, the result of pairwise tests can be presented in the form of letters  \cite{Piepho2004-lt}. 
We use the letters \emph{a} to \emph{z}. 
Every level is assigned a combination of letters, often only a single one. 
If two levels share a letter, the respective test is not significant; that is, H$_A$ from \cref{hyt:wilc_test} cannot be accepted.

\subsubsection{Kendall's Tau Coefficient}

When analysing the results of our experiments in terms of the different $\dataset$, we are interested in the effects of the number of instances $\ngenu$ and the number of attributes $\ndim$. 
Hence, we are interested in the dependency of the mcc score on $\ngenu$ or $\ndim$. 
A common estimate for a monotonic relationship between two random variables is Kendall's Tau ($\tau$) \cite{prokhorov2001kendall}. 
The situation of $\tau = 0$ indicates that there is no dependency, $\tau > 0$ stands for a joint increase, and $\tau < 0$ indicates that an increase in one variable leads to a decrease in the other. 
We make use of the Tau test \cite{prokhorov2001kendall} formalized in \cref{hyt:tau_test} to test whether the estimated $\tau$ is significant.

\begin{hyptest}[Tau Test]
	\label{hyt:tau_test}
	Let $X$ and $Y$ be two random variables with $\tau=\tau_0$. 
	The null and alternative hypothesis of the tau test are
	\begin{equation}
	\text{H}_0 : \tau_0 = 0 \quad \text{and} \quad \text{H}_A : \tau_0 \neq 0.
	\end{equation}
\end{hyptest}
Similarly to the previous test, we have to account for multiple testing. 
We again apply the Holm--Bonferroni method \cite{Holm1979-gc}.

\subsection{Performing the ANOVA}

The factors in our experiments have many levels: $\classifier$ has 3; $\dataset$, 16; $\trainGen$, 11: and $\testGen$, 12. 
We think that this large number of levels and factors makes a full ANOVA with all possible interactions difficult to interpret.
To reduce the number of levels, we use an aggregated type of $\testGen$: the genuine outliers from $\dataset$ ($\trueOuts$) and the median of all generating approaches ($\artOuts$). 
We aggregate the result on all generation approaches since we are not very interested in the detection quality of a single type of artificial outliers. 
A low detection quality could, for example, mean simply that these outliers are easy to detect and not offer any insight into the ability of the specific classifier to detect various types of outliers. 
The results of the ANOVA are displayed in \cref{tab:anova}.

\begin{table}[ht]
	\centering
	\caption{Four-Way Anova from Experiments.} 
	\label{tab:anova}
\begin{tabular}{lrrrr}
  \toprule
Source & \makecell{Sum of \\ Squares} & \makecell{Degrees of \\ Freedom} & F-Value & $\omega^2_P$ \\ 
  \midrule
$\dataset^1$ & 366.82 & 15 & 2319.90 & 0.62 \\ 
  $\trainGen^2$ & 265.90 & 10 & 2522.51 & 0.54 \\ 
  $\text{Inter}(1, 3)$ & 191.31 & 15 & 1209.93 & 0.46 \\ 
  $\text{Inter}(1, 2)$ & 182.84 & 150 & 115.63 & 0.45 \\ 
  $\text{Inter}(1, 2, 3)$ & 166.33 & 150 & 105.19 & 0.43 \\ 
  $\testGen^3$ & 148.82 & 1 & 14117.71 & 0.40 \\ 
  $\text{Inter}(1, 2, 4)$ & 124.99 & 300 & 39.52 & 0.35 \\ 
  $\text{Inter}(2, 4)$ & 106.72 & 20 & 506.20 & 0.32 \\ 
  $\text{Inter}(2, 3)$ & 87.96 & 10 & 834.43 & 0.28 \\ 
  $\text{Inter}(1, 2, 4, 3)$ & 61.52 & 300 & 19.45 & 0.21 \\ 
  $\text{Inter}(1, 4)$ & 44.32 & 30 & 140.13 & 0.17 \\ 
  $\text{Inter}(1, 4, 3)$ & 18.25 & 30 & 57.71 & 0.07 \\ 
  $\text{Inter}(2, 4, 3)$ & 17.54 & 20 & 83.22 & 0.07 \\ 
  $\classifier^4$ & 1.26 & 2 & 59.59 & 0.01 \\ 
  $\text{Inter}(4, 3)$ & 0.45 & 2 & 21.34 & 0.00 \\ 
  Residuals & 211.50 & 20064 &  &  \\ 
   \bottomrule
\end{tabular}

\end{table}

\noindent The \emph{F}-tests for each source yield a highly significant result (\emph{p}-values $< 6 \cdot e^{-10}$). 
Thus, at least one mean within the different levels of each source significantly differs from the other levels. 
We conclude that each source listed in \cref{tab:anova} determines to some extent whether the mcc score of an experiment is high or low on average. 
The significance of all possible interactions means that the main factors influence each other.
For example, the choice of a classifier type has an impact on the generation approach for outliers that results in a high mcc score on average. 
This finding coincides with what \citet{Hastie2009-mu} and~\citet{Steinwart2005-vz} have hypothesized. 
Note that the necessary assumption for ANOVA of standard normal residuals with equal variance is not fully met in our case. 
Thus, the \emph{p}-values of the \emph{F}-test and the $\omega^2_P$ values might not be exact. 
Our subsequent \emph{post hoc} analysis does, however, confirm the tests regarding the main factors.

\subsection{Classifier Comparison}
\label{sec:compare_class}

From the $\omega^2_P$ values in \cref{tab:anova}, we see that the type of classifier ($\classifier$) itself has a rather small impact on the variation of the resulting mcc scores. 
As such, it is not important to explain a high or low mcc score. 
Interestingly, some interactions involving the classifier are ranked much higher, the one of classifier and $\trainGen$, for example. 
Hence, it is more important that the classifier and the generation approach for outliers fit. 
However, we are interested in the exact differences of the types of classifier. \cref{tab:classifier_wilcox} features the results of the respective pairwise Mann--Whitney \emph{U} test.

\begin{table}[ht]
	\centering
	\caption{Comparison of different types of classifier.} 
	\label{tab:classifier_wilcox}
\begin{tabular}{lrrl}
  \toprule
$\classifier$ & Mean & Median & Test Result \\ 
  \midrule
binaryGrid & 0.31 & 0.24 & \phantom{a}b \\ 
  one-class & 0.30 & 0.24 & \phantom{a}b \\ 
  binary & 0.29 & 0.23 & a\phantom{b} \\ 
   \bottomrule
\end{tabular}

\end{table}

\begin{figure}[ht]
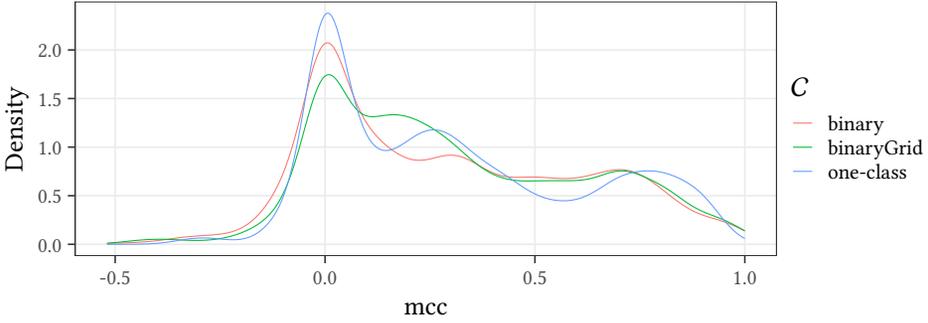

	\centering
	\include{figures/classifier_dens}
	\caption{Density of mcc score with different classifier types.}
	\Description{All three densities have a peak at a mcc of 0 with a density of about 2. For a mcc lower than 0 all densities rapidly decreases until they reach zero at about -0.5. For positive mcc values the densities gradually but quite wiggly decrease until they are close to zero at a mcc value of 1. The densities of the one-class and binaryGrid classifier concentrate their mass slightly more on positive mcc values than the binary classifier.}
	\label{fig:classifier_dens}
\end{figure}

\noindent We see that the one-class and binaryGrid classifier have the same group letter. 
Hence, there seem to be few reasons to prefer one over the other. 
The binary classifier is in its own group, indicating a significant difference from the former two. The mean and median mcc score are lowest. However, the difference is tiny.
This results differ somewhat from the results of the small study in \cite{Davenport2006-ck}. 
The results in \cite{Davenport2006-ck} suggest that the \emph{casting tasks} use case (binary/binaryGrid classifier) is preferable.
To further elaborate on the difference in distributions of the classifier types, we visualize the estimated probability density of the mcc score for the different classifier types in \cref{fig:classifier_dens}. 
The figure visually supports that the differences are not great, but the binaryGrid or one-class classifier are more likely to result in high mcc scores.

\subsection{Comparison of Generation Approaches}

\begin{figure}[ht]
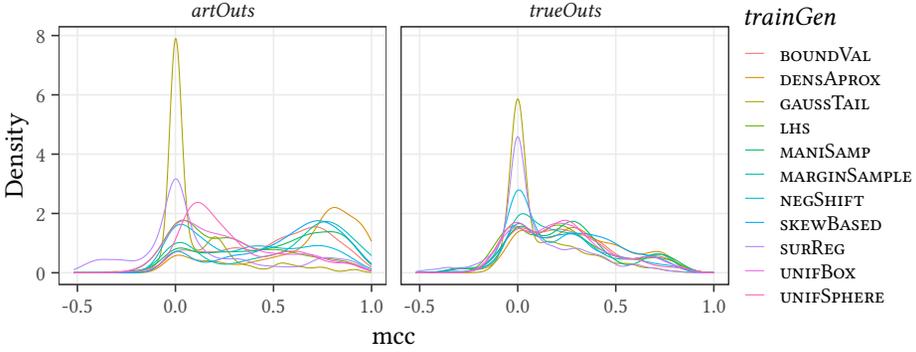

	\centering
	\include{figures/testgen_dens}%
	\caption{Density of mcc score with different generation approaches.}
	\Description{Two density plots are shown, one for artificial outliers and one for true outliers. In each the densities regarding the achieved mcc value for a certain approach for generating artificial outlier in the training phase are given. Densities in the plot for artificial outliers often have a peak at a mcc value of 0 and about 0.75. The ones in the plot for true outliers have a peak at 0, 0.25 and 0.75. In general the densities in the plot regarding true outliers are more distributed. I.e., less mass is concentrated.}
	\label{fig:testGen_dens}
\end{figure}

The $\omega^2_P$ values in \cref{tab:anova} for the two factors that relate to generation approaches, $\trainGen$ and $\testGen$, are quite high --- for $\trainGen$, in particular. 
Thus, they account for a large part of the variation in the mcc score. 
The interpretation of the generation approaches for $\trainGen$ and $\testGen$ differ greatly. 
We start with the results in terms of $\trainGen$ and then analyze the results regarding $\testGen$. 
\cref{fig:testGen_dens} displays the mcc score probability density regarding the levels of $\trainGen$ and $\testGen$. 

\subsubsection{Artificial Outliers for Training}

The $\omega^2_P$ of $\trainGen$ is the second highest in \cref{tab:anova}. 
Thus, certain significant differences in detection quality appear when using different generation approaches to train the classifiers. 
\cref{tab:trainGen_wilcox} lists the results of the pairwise Mann--Whitney \emph{U} tests. 
The approaches $\boundVal$ and $\maniSamp$ form a group, and $\unifBox$, $\lhs$, $\unifSphere$ and $\negShift$ form one as well. 
The elements of the second group, in particular, are conceptually quite similar. 
For example, three of the four approaches spread outliers uniformly. 
The approaches with the highest mean and median are $\densAprox$ and $\skewBased$. 
Both try to generate outliers similar to the genuine instances. 
The $\densAprox$ approach does this quite literally. 
Hence, we conclude that this is a generally useful approach to generate outliers that classifiers are trained with. 
Somewhat contradictory to this conclusion, however, is that the $\surReg$ approach, which also generates outliers similar to genuine instances, shares the lowest mean and median with $\gaussTail$. 
From \cref{fig:testGen_dens}, we see that both approaches often result in an mcc score close to zero. 
Hence, training the classifier using artificial outliers generated by $\surReg$ or $\gaussTail$ seems to result in a rather low detection quality. 
For the $\gaussTail$ approach, we think that this is because the generated outliers are too far from genuine instances to be interesting (cf.\ \cref{exp:interesting_place}). 
The distribution of outliers generated by the $\surReg$ approach is heavily influenced by the attribute bounds. 
This influence might lead to an uneven coverage of the instance space around genuine instances. 
It might also be that instances generated with $\surReg$ are too close to genuine ones to be interesting, which would explain the low mcc score when outliers generated with $\surReg$ are used to test a classifier (cf.\ \cref{tab:testGen_wilcox}).

\begin{table}[ht]
	\centering
	\caption{Comparison of artificial outliers for training.} 
	\label{tab:trainGen_wilcox}
\begin{tabular}{lrrl}
  \toprule
$\trainGen$ & Mean & Median & Test Result \\ 
  \midrule
$\densAprox$ & 0.47 & 0.49 & \phantom{a}b\phantom{c}\phantom{d}\phantom{e}\phantom{f} \\ 
  $\skewBased$ & 0.41 & 0.42 & \phantom{a}\phantom{b}\phantom{c}\phantom{d}e\phantom{f} \\ 
  $\marginSample$ & 0.39 & 0.34 & a\phantom{b}\phantom{c}\phantom{d}e\phantom{f} \\ 
  $\boundVal$ & 0.37 & 0.35 & a\phantom{b}\phantom{c}\phantom{d}\phantom{e}\phantom{f} \\ 
  $\maniSamp$ & 0.37 & 0.33 & a\phantom{b}\phantom{c}\phantom{d}\phantom{e}\phantom{f} \\ 
  $\negShift$ & 0.28 & 0.22 & \phantom{a}\phantom{b}\phantom{c}d\phantom{e}\phantom{f} \\ 
  $\lhs$ & 0.26 & 0.21 & \phantom{a}\phantom{b}\phantom{c}d\phantom{e}\phantom{f} \\ 
  $\unifBox$ & 0.26 & 0.21 & \phantom{a}\phantom{b}\phantom{c}d\phantom{e}\phantom{f} \\ 
  $\unifSphere$ & 0.25 & 0.21 & \phantom{a}\phantom{b}\phantom{c}d\phantom{e}\phantom{f} \\ 
  $\gaussTail$ & 0.11 & 0.00 & \phantom{a}\phantom{b}c\phantom{d}\phantom{e}\phantom{f} \\ 
  $\surReg$ & 0.11 & 0.00 & \phantom{a}\phantom{b}\phantom{c}\phantom{d}\phantom{e}f \\ 
   \bottomrule
\end{tabular}

\end{table}

\subsubsection{Artificial Outliers for Testing}
\label{sec:art_outs_test}

Since we aggregate all artificial outlier-generation approaches for the ANOVA, $\trainGen$ has only two values: $\artOuts$ and $\trueOuts$. 
Hence, we can conclude immediately that there is a significant difference in the mean mcc score of the two (cf.\ \cref{tab:testGen_wilcox}). 
This implies that there is quite a gap between the quality we assign to a detection method when we evaluate it with the artificial outlier types presented or the labeled ground truth outliers from the benchmark data sets. 
This is also clearly visible in \cref{fig:testGen_dens}. 
We hypothesize that this is mainly because most artificial outliers are much simpler to identify as such. 
For example, artificial outliers generated with the $\unifBox$ approach tend to be quite far from genuine instances and are hence trivial to classify as outlying. 
Thus, when using artificial outliers to assess the quality of an outlier-detection method, one should consider how difficult the generated outliers  generally are to detect. 
We also think that using several types of outliers (i.e., generated with different approaches) offers much useful insight into the performance of outlier-detection methods.

\begin{table}[ht]
	\centering
	\caption{Comparison of artificial outliers for testing.} 
	\label{tab:testGen_wilcox}
\begin{tabular}{lrrl}
  \toprule
$\testGen$ & Mean & Median & Test Result \\ 
  \midrule
$\gaussTail$ & 0.68 & 0.88 & \phantom{a}\phantom{b}c\phantom{d}\phantom{e}\phantom{f}\phantom{g}\phantom{h}\phantom{i} \\ 
  $\unifBox$ & 0.65 & 0.84 & \phantom{a}\phantom{b}\phantom{c}\phantom{d}\phantom{e}\phantom{f}\phantom{g}\phantom{h}i \\ 
  $\lhs$ & 0.64 & 0.82 & \phantom{a}\phantom{b}\phantom{c}d\phantom{e}\phantom{f}\phantom{g}\phantom{h}\phantom{i} \\ 
  $\unifSphere$ & 0.62 & 0.81 & \phantom{a}\phantom{b}\phantom{c}d\phantom{e}\phantom{f}\phantom{g}\phantom{h}\phantom{i} \\ 
  $\densAprox$ & 0.29 & 0.21 & a\phantom{b}\phantom{c}\phantom{d}\phantom{e}\phantom{f}\phantom{g}\phantom{h}\phantom{i} \\ 
  $\boundVal$ & 0.29 & 0.28 & ab\phantom{c}\phantom{d}\phantom{e}\phantom{f}\phantom{g}\phantom{h}\phantom{i} \\ 
  $\maniSamp$ & 0.29 & 0.19 & \phantom{a}b\phantom{c}\phantom{d}\phantom{e}\phantom{f}\phantom{g}\phantom{h}\phantom{i} \\ 
  $\skewBased$ & 0.28 & 0.09 & \phantom{a}\phantom{b}\phantom{c}\phantom{d}e\phantom{f}g\phantom{h}\phantom{i} \\ 
  $\textbf{\trueOuts}$ & 0.21 & 0.18 & \phantom{a}\phantom{b}\phantom{c}\phantom{d}\phantom{e}\phantom{f}g\phantom{h}\phantom{i} \\ 
  $\negShift$ & 0.21 & 0.06 & \phantom{a}\phantom{b}\phantom{c}\phantom{d}\phantom{e}f\phantom{g}\phantom{h}\phantom{i} \\ 
  $\marginSample$ & 0.20 & 0.12 & \phantom{a}\phantom{b}\phantom{c}\phantom{d}e\phantom{f}\phantom{g}\phantom{h}\phantom{i} \\ 
  $\surReg$ & 0.05 & 0.00 & \phantom{a}\phantom{b}\phantom{c}\phantom{d}\phantom{e}\phantom{f}\phantom{g}h\phantom{i} \\ 
  $\textbf{\artOuts}$ & 0.38 & 0.35 &  \\ 
   \bottomrule
\end{tabular}

\end{table}

Although we have aggregated the generation approaches used for testing the classifiers when performing the ANOVA, we are nevertheless interested in the differences of the generation approaches presented. 
Note that a high mcc value here means that the outliers generated are generally easy to detect. 
\cref{tab:testGen_wilcox} lists the results of corresponding pairwise  Mann--Whitney \emph{U} tests. 
The mean and median of the aggregated version $\artOuts$ are also listed as references. 
Approaches used to test the classifiers do not group much; only $\lhs$ and $\unifSphere$ are in one group.
We also observe that the ranking of generation approaches in \cref{tab:testGen_wilcox} is to some extent inverse to the one in \cref{tab:trainGen_wilcox}. 
For example, $\gaussTail$ is listed first in \cref{tab:testGen_wilcox} but second-to-last in \cref{tab:trainGen_wilcox}.
We think that this listing further supports our previous hypothesis on the successful generation approaches to train a classifier. 
Training a classifier with outliers that are somewhat similar to the genuine instances, and hence more difficult to detect, results in a better detection method.

\subsection{Data Characteristics}

From the ANOVA results in \cref{tab:anova}, we see that the underlying data set $\dataset$ is quite important to determine if the mcc score of an experiment is rather high or low on average. 
A more detailed analysis regarding the effect of $\dataset$ can be found in \cref{tab:dataset_wilcox} with the pairwise Mann--Whitney \emph{U} test. 
Some data sets share a letter and hence do not allow for the acceptance of the alternative hypothesis that one is stochastically larger than the other one, but most are in different groups.

\begin{table}[ht]
	\centering
	\caption{Comparison of different data sets.} 
	\label{tab:dataset_wilcox}
\begin{tabular}{lrrrrl}
  \toprule
$\dataset$ & $\ngenu$ & $\ndim$ & Mean & Median & Test Result \\ 
  \midrule
PageBlocks & 1000 & 10 & 0.61 & 0.67 & \phantom{a}\phantom{b}\phantom{c}\phantom{d}\phantom{e}\phantom{f}\phantom{g}\phantom{h}\phantom{i}j\phantom{k}\phantom{l} \\ 
  Ionosphere & 351 & 32 & 0.48 & 0.62 & \phantom{a}\phantom{b}\phantom{c}\phantom{d}\phantom{e}\phantom{f}g\phantom{h}\phantom{i}\phantom{j}\phantom{k}\phantom{l} \\ 
  Stamps & 340 & 9 & 0.42 & 0.43 & \phantom{a}\phantom{b}\phantom{c}\phantom{d}e\phantom{f}\phantom{g}\phantom{h}\phantom{i}\phantom{j}\phantom{k}\phantom{l} \\ 
  Glass & 214 & 7 & 0.42 & 0.39 & \phantom{a}\phantom{b}\phantom{c}\phantom{d}e\phantom{f}\phantom{g}\phantom{h}\phantom{i}\phantom{j}\phantom{k}\phantom{l} \\ 
  KDDCup99 & 1000 & 40 & 0.38 & 0.31 & \phantom{a}\phantom{b}\phantom{c}\phantom{d}\phantom{e}\phantom{f}\phantom{g}h\phantom{i}\phantom{j}\phantom{k}\phantom{l} \\ 
  Wilt & 1000 & 5 & 0.33 & 0.21 & \phantom{a}bc\phantom{d}\phantom{e}\phantom{f}\phantom{g}\phantom{h}\phantom{i}\phantom{j}\phantom{k}l \\ 
  Cardiotocography & 1000 & 21 & 0.32 & 0.28 & \phantom{a}\phantom{b}\phantom{c}d\phantom{e}\phantom{f}\phantom{g}\phantom{h}\phantom{i}\phantom{j}\phantom{k}\phantom{l} \\ 
  Pima & 768 & 8 & 0.29 & 0.21 & \phantom{a}\phantom{b}cd\phantom{e}\phantom{f}\phantom{g}\phantom{h}\phantom{i}\phantom{j}\phantom{k}\phantom{l} \\ 
  Annthyroid & 1000 & 21 & 0.28 & 0.25 & \phantom{a}\phantom{b}c\phantom{d}\phantom{e}\phantom{f}\phantom{g}\phantom{h}\phantom{i}\phantom{j}\phantom{k}\phantom{l} \\ 
  SpamBase & 1000 & 57 & 0.23 & 0.18 & \phantom{a}\phantom{b}\phantom{c}\phantom{d}\phantom{e}\phantom{f}\phantom{g}\phantom{h}\phantom{i}\phantom{j}kl \\ 
  ALOI & 1000 & 27 & 0.22 & 0.06 & ab\phantom{c}\phantom{d}\phantom{e}\phantom{f}\phantom{g}\phantom{h}\phantom{i}\phantom{j}\phantom{k}\phantom{l} \\ 
  Parkinson & 195 & 22 & 0.22 & 0.22 & \phantom{a}\phantom{b}\phantom{c}\phantom{d}\phantom{e}\phantom{f}\phantom{g}\phantom{h}\phantom{i}\phantom{j}k\phantom{l} \\ 
  WPBC & 198 & 33 & 0.16 & 0.08 & \phantom{a}\phantom{b}\phantom{c}\phantom{d}\phantom{e}f\phantom{g}\phantom{h}i\phantom{j}\phantom{k}\phantom{l} \\ 
  musk2 & 1000 & 166 & 0.16 & 0.00 & \phantom{a}\phantom{b}\phantom{c}\phantom{d}\phantom{e}\phantom{f}\phantom{g}\phantom{h}i\phantom{j}\phantom{k}\phantom{l} \\ 
  HeartDisease & 270 & 13 & 0.15 & 0.13 & a\phantom{b}\phantom{c}\phantom{d}\phantom{e}\phantom{f}\phantom{g}\phantom{h}\phantom{i}\phantom{j}\phantom{k}\phantom{l} \\ 
  Hepatitis & 80 & 19 & 0.13 & 0.09 & \phantom{a}\phantom{b}\phantom{c}\phantom{d}\phantom{e}f\phantom{g}\phantom{h}\phantom{i}\phantom{j}\phantom{k}\phantom{l} \\ 
   \bottomrule
\end{tabular}

\end{table}

We hypothesize that most of the difference in detection quality is due to the distribution of the different data sets. 
However, the numbers of genuine instances or attributes also have an effect. 
We also think that the generation approach used to train the classifier has a strong influence on this effect. 
Thus, for each level of $\trainGen$, we estimate $\tau$ between the mcc and the number of attributes $\ndim$ as well as the number of genuine instances $\ngenu$. 
For each $\tau$ we also perform a tau test. 
The results are in displayed \cref{tab:cor_test}, and those with a significant $\tau$ (\emph{p}-value $< 0.05$) are in bold. 
The \emph{p}-value is abbreviated as p$_d$ or p$_n$.

\begin{table}[ht]
	\centering
	\caption{Correlation with $\ndim$ and $\ngenu$.} 
	\label{tab:cor_test}
\begin{tabular}{llrlr}
  \toprule
$\trainGen$ & $\hat{\tau}_\ndim$ & p$_\ndim$ & $\hat{\tau}_\ngenu$ & p$_\ngenu$ \\ 
  \midrule
$\boundVal$ & \textbf{-0.13} & 0.00 & 0.03 & 0.21 \\ 
  $\densAprox$ & \textbf{0.05} & 0.00 & \textbf{0.21} & 0.00 \\ 
  $\gaussTail$ & \textbf{-0.07} & 0.00 & \textbf{0.05} & 0.03 \\ 
  $\lhs$ & \textbf{-0.24} & 0.00 & -0.04 & 0.08 \\ 
  $\maniSamp$ & -0.01 & 1.00 & \textbf{0.17} & 0.00 \\ 
  $\marginSample$ & 0.01 & 1.00 & \textbf{0.10} & 0.00 \\ 
  $\negShift$ & \textbf{-0.23} & 0.00 & 0.01 & 0.49 \\ 
  $\skewBased$ & -0.03 & 0.16 & \textbf{0.17} & 0.00 \\ 
  $\surReg$ & \textbf{-0.06} & 0.00 & \textbf{0.28} & 0.00 \\ 
  $\unifBox$ & \textbf{-0.28} & 0.00 & -0.04 & 0.07 \\ 
  $\unifSphere$ & \textbf{-0.26} & 0.00 & -0.03 & 0.11 \\ 
   \bottomrule
\end{tabular}

\end{table}

\noindent The $\tau_d$ tend to be negative, indicating a decreasing effect on the mcc score for an increasing number of attributes. 
This relation seems to be particularly strong with the $\unifBox$ approach. 
Most of the $\tau_n$ are significant and positive. 
Hence, if a higher number of instances has an effect on the resulting mcc score at all, the effect is usually a positive one.

\subsection{Summary of Experiments}

\begin{figure}[ht]
	\centering
	\begin{tikzpicture}[
	basic/.style  = {draw, text width=6cm, rectangle, rounded corners=2pt, thin, align=center},
	edge from parent/.style={->,draw},
	>=latex]
	
	\node[basic] {Type of Classifier}
	child {node[basic] (c1) {Generation Approach for Training}
		child {node[basic] (c11) {Generation Approach for Testing}}
	};
	\end{tikzpicture}
	\caption{Process to choose outlier-generation approach.}
	\Description{The figure shows the three sequential steps we propose for choosing a outlier generation approaches. First choose a type of classifier, then a generation approach for training and finally a generation approach for testing.}
	\label{fig:choose_gen_approach}
\end{figure}
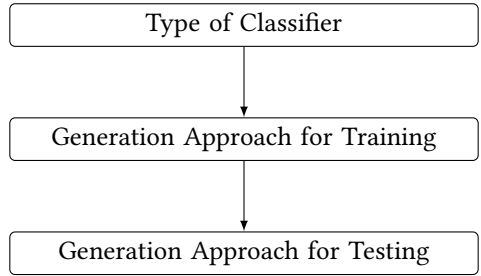

A core insight from our study is that there are huge differences in the outlier-detection quality for different generation approaches and data sets. 
When used to train a classifier, the overall best performing approach has been $\densAprox$, with a median mcc of 0.49. 
The worst ones have been $\gaussTail$ and $\surReg$, both with a median mcc of 0. 
This result is comparable to those obtained by random guesses.
The data sets form only few groups with no significant difference in terms of the overall mcc to other data sets. 
However, there are significant differences between the groups. 
For example, with the Ionosphere data set from the group with letter g, the median mcc is 0.62, while it is only 0.08 with the WPBC from the group with letters fi. 
All interactions between the main factors are significant. 
Thus, the choice of a generation approach in a specific scenario cannot be reduced to, for example, \enquote{$\densAprox$ performs best}. 
Depending on the classifier a scenario requires or on the data set given by the scenario, different approaches might be suitable. 
This realization has motivated us to propose a three-step process, displayed in \cref{fig:choose_gen_approach}, in order to choose a generation approach in a specific scenario. 
The steps are based on the results of our experiments. 
They help us to make the necessary decisions when the goal is to detect outliers with the help of artificial outliers.

\subsubsection*{Step 1: Type of Classifier}

When the ultimate goal is a method to detect outliers, any of the use cases \emph{casting tasks} and \emph{one-class tuning} is applicable. 
In our experiments, we have found that the \emph{one-class} or \emph{binaryGrid} classifier yield similar mcc scores.
However, in a specific scenario the huge variety of binary classifiers available for the \emph{casting task} can be advantageous. 
One example is when the outlier-detection result should be easily interpretable.
A decision tree might then be a better fit than a one-class SVM. 

\subsubsection*{Step 2: Generation Approach for Training}

One can now check for outlier generating approaches that are more suitable for the classifier chosen in Step 1. 
The experimental results displayed in this survey may be very helpful, but are not necessarily sufficient to this end. 
Observe, however, that this article does not explicitly feature the result in every useful representation, to ensure that this survey still has a reasonable length. 
The full results are available, however, in combination with our code. 
Others can also use the code to test further combinations; this may be particularly useful when new types of classifiers become available. 
In addition to the type of classifier used, the data set of the scenario is of importance. 
One can, for example, check whether this data set is similar to one of the data sets from our experiments, or if the number of attributes and genuine instances is high or low. 
Depending on these two factors (classifier and data set), one can then choose the best-suited outlier-generation approach to train the classifier.

\subsubsection*{Step 3: Generation Approach for Testing}

We have seen in \cref{sec:art_outs_test} that one needs to be careful when assessing the quality of outlier detection using artificial outliers. 
We think that this assessment, nevertheless, offers useful insight into the outlier-detection quality. 
In a real-world scenario, there might be some knowledge of potential genuine outliers available. 
Consider a system administrator who has a rough idea of how possible outliers might be distributed. Suppose further that this distribution is somewhat similar to the one of artificial outliers generated by the $\negShift$ approach. 
Detection quality in terms of artificial outliers generated with $\negShift$ is then clearly a good estimate for the detection quality of genuine instances. 
However, if there is no such knowledge, which might be the much more likely case, we conclude that one of the quite general and uninformative approaches to artificial outliers, like $\lhs$, is well-suited. 
To gain a better feel for which types of outliers are and are not well detected, a quality assessment using a variety of the generation approaches described might also be suitable. 
However, we leave a systematic study of this idea into future work because this is not straightforward at all, and it goes well beyond the scope of this survey.

\section{Conclusions}
\label{sec:conclusion}

This section presents a summary of our work and its limitations. 
Promising directions for future research are discussed as well.

\subsection{Summary}

By definition, outliers are instances that are rarely observed in reality, so it is difficult to learn anything with them. 
To compensate for this shortage of data, various approaches to generate artificial outliers have been proposed. 
This article is a survey of such approaches. 
As a first step, we have connected the field of artificial outliers to other research fields.     
This step allows us to narrow down the field of artificial outliers somewhat. 
The generation approaches described next represent rather different ways to generate artificial outliers. 
They form separate groups, depending on the similarity of the generated instances to genuine ones or on the general generation concept. 
All this results in the general perspective on artificial outliers we aimed at.

Depending on the use case for the artificial outliers, different generation approaches might yield \emph{interesting} artificial outliers. 
Our experiments confirm the hypothesis of some authors that, for the \emph{one-class tuning} or \emph{casting task} use case, artificial outliers similar to genuine instances seem to be interesting.
The experiments also confirm that this interestingness heavily depends on the setting (e.g., the data set used). 
In terms of the use cases themselves, our experiments suggest that there is no distinctive differences in outlier detection performance.
Analysing the effect of some data set characteristic with different generation approaches confirms that these can heavily influence the outlier-detection performance.

To this end, we have also developed a decision process, building on the results of our experiments, that guides the choice of a good generation approach. 
In other words, the process targets at finding a generation approach that yields high outlier-detection quality.

\subsection{Limitations}

This study focuses on the description, categorization, and comparison of the various existing generation approaches for artificial outliers. 
We have not proposed any new generation approach, but only compared the existing ones, mainly quantitatively. 
Beyond the detail that is necessary to this end, we have not yet carried out any further investigation of the behavior of the different approaches and see this as future work. 
Besides this, we have not actively questioned the value and purpose of artificial outliers in addition to what others have already observed. 
We also do not use the generation approaches to benchmark outlier-detection algorithms, because this would have exceeded the scope of this study by much.
Nevertheless, our study marks a very good starting point for anyone interested in the topic of artificial outliers. 
We do show where the spectrum of the existing approaches is ranging, how well the approaches perform in specific settings, what is currently achievable in terms of outlier-generation quality and uncover areas with potential for future work.

\subsection{Future Research Directions}

Our study reveals that there are numerous questions regarding artificial outliers that require attention in the future. 
One applies to the limit of the similarity of artificial outliers and genuine instances mentioned earlier. 
If the distribution of artificial outliers and the one of the genuine instances completely fall together, there is nothing to gain from the artificial outliers. 
Thus, investigating when artificial outliers are \enquote{too similar} and no longer useful is an interesting future research challenge.  
Another question regards the effect of the number of artificial outliers. 
Although some studies have recognized its importance (e.g., in \cite{Hastie2009-mu}), the issue is often not explicitly addressed. 
However, from the connection to  generative models (\cref{sec:gen_models}), we see that that number has a strong effect on the decision of whether an instance is an outlier. 
We also find it worth investigating how the various methods to filter artificial instances interact with the generation approaches. 
That is, can there be guidelines on when to use which filter? 
Another interesting future research direction is the connection of the  approaches presented to methods extending a set of genuine outliers. 
It could well be that novel approaches can be developed integrating ideas from both fields: generation with and without genuine outliers. 
Finally, further assessments of artificial outliers as a means of evaluating outlier detection would be useful. 
To illustrate, one way to do so could be to develop a framework which systematically tests the outlier-detection results with diverse types of artificial outliers. 
This might improve the evaluation of outlier-detection methods by much. 

Our study has featured a great variety of approaches for the generation of artificial outliers. 
The experimental study we have conducted is a basis for the decision-making process towards a good outlier-generation approach. 
The study also has revealed many possible future research directions. 
As such, this study is likely to support individuals from diverse fields when developing advanced approaches for the generation of artificial outliers. 

\begin{acks}
	This work has been supported by the \grantsponsor{dfg}{German Research Foundation (DFG)}{} as part of the \grantnum{dfg}{Research Training Group GRK 2153: Energy Status Data -- Informatics Methods for its Collection, Analysis and Exploitation}.
\end{acks}

\bibliographystyle{ACM-Reference-Format}
\bibliography{bib/paper}
	
\end{document}